\documentclass[lettersize,journal]{IEEEtran}
\usepackage{amsmath,amsfonts}
\usepackage{algorithm}
\usepackage{array}
\usepackage[caption=false,font=normalsize,labelfont=sf,textfont=sf]{subfig}
\usepackage{textcomp}
\usepackage{stfloats}
\usepackage{url}
\usepackage{verbatim}
\usepackage{graphicx}
\usepackage{cite}
\usepackage{graphicx}
\usepackage{multirow}
\usepackage{caption}
\usepackage{xcolor}
\usepackage{xcolor}
\usepackage[colorlinks = true,
            linkcolor = blue,
            urlcolor  = blue,
            citecolor = blue,
            anchorcolor = blue]{hyperref}
\usepackage{algorithm}
\usepackage{algpseudocode}
\usepackage{todonotes}
\usepackage{soul}
\usepackage{multirow}
\usepackage{booktabs}
\usepackage{lipsum}
\usepackage{pifont}
\newcommand{\cmark}{\ding{51}}%
\newcommand{\xmark}{\ding{55}}%
\usepackage[switch]{lineno} 
\usepackage[normalem]{ulem}
\begin{document}
\title{FedDCT: Federated Learning of Large Convolutional Neural Networks on Resource Constrained Devices using Divide and Collaborative Training}

\author{Quan Nguyen, Hieu H. Pham$^{*}$, \textit{Member, IEEE}, Kok-Seng Wong, \textit{Member, IEEE}, Phi Le Nguyen, Truong Thao Nguyen and Minh N. Do, \textit{Fellow, IEEE}

\IEEEcompsocitemizethanks{
\IEEEcompsocthanksitem Quan Nguyen, Kok-Seng Wong, Hieu H. Pham and Minh Do are with the VinUni-Illinois Smart Health Center, VinUniversity, Hanoi, Vietnam.\protect\\
E-mail: \{quan.nm,hieu.ph,wong.ks,minh.do@vinuni.edu.vn\}
\IEEEcompsocthanksitem Quan Nguyen is also with Department of Informatics, Technical University of Munich, Munich, Germany.\protect\\
E-mail: quan.nguyen@tum.de
\IEEEcompsocthanksitem Hieu H. Pham, Kok-Seng Wong, and Minh Do are also with the College of Engineering and Computer Science, VinUniversity, Hanoi, Vietnam.
\IEEEcompsocthanksitem Phi Le is with the School of Information and Communication Technology, Hanoi University of Science and Technology, Hanoi, Vietnam.\\
Email: {lenp@soict.hust.edu.vn}
\IEEEcompsocthanksitem Truong Thao Nguyen is with the National Institute of Advanced Industrial Science and Technology (AIST), Tokyo, Japan.\\
Email: {nguyen.truong@aist.go.jp}
\IEEEcompsocthanksitem Minh N. Do is also with the Department of Electrical and Computer Engineering, University of Illinois at Urbana-Champaign, Illinois, USA.\\
Email: {minhdo@illinois.edu  \\ $*$ Corresponding author: \textcolor{blue}{hieu.ph@vinuni.edu.vn} (Hieu H. Pham) }
}}

\markboth{}%
{Shell \MakeLowercase{\textit{et al.}}: A Sample Article Using IEEEtran.cls for IEEE Journals}

\bstctlcite{IEEEexample:BSTcontrol} 
\maketitle

\begin{abstract}
In Federated Learning (FL), the size of local models matters. On the one hand, it is logical to use large-capacity neural networks in pursuit of high performance. On the other hand, deep convolutional neural networks (CNNs) are exceedingly parameter-hungry, which makes memory a significant bottleneck when training large-scale CNNs on hardware-constrained devices such as smartphones or wearables sensors. Current state-of-the-art (SOTA) FL approaches either only test their convergence properties on tiny CNNs with inferior accuracy or assume clients have the adequate processing power to train large models, which remains a formidable obstacle in actual practice. 
To overcome these issues, we introduce FedDCT, a novel distributed learning paradigm that enables the usage of large, high-performance CNNs on resource-limited edge devices. As opposed to traditional FL approaches, which require each client to train the full-size neural network independently during each training round, the proposed FedDCT allows a cluster of several clients to collaboratively train a large deep learning model by dividing it into an ensemble of several small sub-models and train them on multiple devices in parallel while maintaining privacy. In this collaborative training process, clients from the same cluster can also learn from each other, further improving their ensemble performance. In the aggregation stage, the server takes a weighted average of all the ensemble models trained by all the clusters. FedDCT reduces the memory requirements and allows low-end devices to participate in FL. We empirically conduct extensive experiments on standardized datasets, including CIFAR-10, CIFAR-100, and two real-world medical datasets HAM10000 and VAIPE. Experimental results show that FedDCT outperforms a set of current SOTA FL methods with interesting convergence behaviors. Furthermore, compared to other existing approaches, FedDCT achieves higher accuracy and substantially reduces the number of communication rounds (with $4-8$ times fewer memory requirements) to achieve the desired accuracy on the testing dataset without incurring any extra training cost on the server side. 

\end{abstract}


\section{Introduction}
\begin{figure}[]
  \centering
  \includegraphics[width=\linewidth]{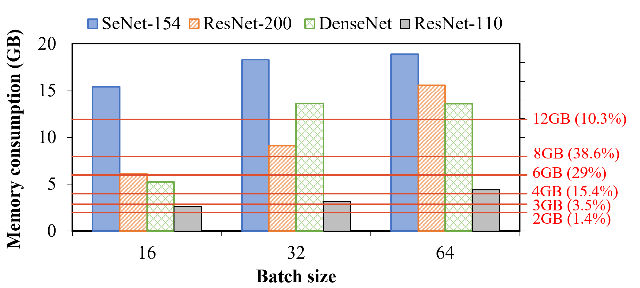}
  \caption{Memory consumption of training SetNet-154~\cite{Hu_2018_CVPR}, ResNet-200~\cite{he2016deep} on ImageNet dataset~\cite{deng2009imagenet}, DensetNet~\cite{Huang_2017_CVPR} on CIFAR-100 dataset and ResNet-110~\cite{he2016deep} on CIFAR-10 dataset~\cite{cifar} with different batch sizes. All implementations are in PyTorch (version 1.10.2) ~\cite{NEURIPS2019_9015}. The red lines indicate the common memory capacities of mobile devices with their relative amount of devices in percentage ~\cite{MobileDevices}.}
  \label{fig:mem}
\end{figure} 
\IEEEPARstart{F}{ederated} learning (FL)\cite{kairouz2021advances} has emerged as a state-of-the-art machine learning (ML) paradigm that allows clients to cooperatively learn a shared model in a decentralized fashion where no private dataset is sent to a central repository. Specifically, the data for the ML tasks are acquired and processed locally at the edge nodes, and only the updated ML parameters are transmitted to the server for aggregation purposes~\cite{bonawitz2019towards,yang2019federated}. FL has been successfully deployed in many industries, including healthcare, telecommunications, IoT, manufacturing, and surveillance systems~\cite{li2020federated,lavaur2022evolution,subramanya2021centralized, li2021cvc}. Given the broad applications of FL, guaranteeing that such a cooperative learning process is reliable is becoming essential.

Despite significant recent milestones in FL~\cite{khan2021federated}, several fundamental challenges are yet to be addressed \cite{li2020federated2}. For instance, edge nodes often have substantial constraints on their resources, such as memory, computation power, communication, and energy, severely limiting their capability of locally training large models. 
The FL algorithm designs often assume all clients can train the entire ML model independently. However, the dedicated memory of edge devices can be insufficient to store all weights and/or intermediate states of large deep learning models during training\cite{sohoni2019low}. Fig.~\ref{fig:mem} compares the memory consumption during the training of deep neural networks of different model sizes with the memory capacity of current mobile devices. Consequently, the availability of the on-device memory constraints the size of the neural networks model in training \cite{dunnmon2019assessment,yi2017optimizing}. 
Typically, the model size is reduced, and the input is similarly reduced in dimension, potentially destroying information during training. \cite{liu2016ssd}. As data collection systems continue to advance, the dimensionality of real-world datasets continues to grow, leading to an ever more pressing need for the ability to train larger and more complex models in FL\cite{hara2017learning,qiu2017learning}. This necessitates the development of more memory-efficient
training algorithms. However, the majority of state-of-the-art techniques designed to reduce inference memory \cite{han2015deep,zhu2017prune} are inapplicable in training memory reduction, as they often require first training a full-size model that is later compressed. The techniques to reduce training memory used in centralized training at a data center/supercomputer are also not applicable in FL. For example, the out-of-core~\cite{KARMA} method moves data out/into extra memory of the CPU from the GPU's memory. Gradient checkpointing~\cite{Chen_Mem,NEURIPS2019_e555ebe0,MLSYS2020_084b6fbb,sohoni2019low} proposes to redundantly recompute the intermediate states of the forward propagation in the backward propagation instead of storing it. Such extra memory and increased computing requirement can not be fulfilled by the edge devices. As a result, FL algorithms are still restricted to using lightweight models or low-bit training. \cite{thapa2022splitfed,he2020group}.
Other Distributed Collaborative Machine Learning (DCML) methods either require significant training time overhead (e.g., Split Learning \cite{poirot2019split}) or are highly dependent on the computing resources on the server side (e.g., SplitFed \cite{thapa2022splitfed,he2020group}).

To this end, we propose FedDCT (\textit{Federated Divide and Collaborative Training}), a novel FL architecture that allows for DCML with considerably lower client memory requirements over traditional FL while introducing no additional training cost on the central server.  FedDCT enables a cluster of several clients to cooperatively train a large deep learning model, in contrast to standard FL, which demands that each client train the full-size neural network independently during each training round. Specifically, FedDCT (channel-wise) divides the single large global model into several small sub-models\footnote{In this paper, we use terms ``sub-model" and ``sub-network", interchangeably.} regarding its parameters and regularization components. The ensemble model of these sub-models is now the new global model. Each client in a cluster will train the deep ensemble network's sub-models up to a certain ''cut layer" during the training phase using their private dataset. The outputs at the cut layers of the sub-models are sent to other clients in the cluster which completes forward propagation without looking at raw data from the data-holding client. Similarly, these clients perform backpropagation until the cut layer and the gradients at the cut layers (and only these gradients) are sent back to the data-owning client to finish the rest of backpropagation. Together clients in a cluster train the full network. To boost network variety and improve ensemble performance, we also train the sub-networks using various viewpoints of the same data\cite{qiao2018deep,li2018research}. This is implemented by applying different data augmentation in practice. We also add Jensen Shannon divergence \cite{menendez1997jensen} among all predictions of sub-networks. From the predicted posterior probability distributions of its peers, one network can gain insightful knowledge about intrinsic object structure information\cite{hinton2015distilling,zhang2018deep}. In the model aggregation stage, the server combines corresponding sub-models to create an ensemble model for each cluster and aggregates the ensembled models from all clusters to obtain an updated global ensemble model. This process allows for collaborative training of a large global model with multiple low-computing clients in FL. Overall, FedDCT is beneficial for resource-constrained environments where training the large model is not feasible.\newline
\renewcommand{\arraystretch}{1.1}
\begin{table*}[]
\caption{Recent progress with low-memory training targeting to different memory categories: \textbf{M}-Model, \textbf{O}-Optimizer, \textbf{A}-Activation. $\downarrow$: reducing memory/computation; $\uparrow$: need extra overhead; -: nothing changed; $*$: not specify.}
\label{table:overview}
\centering
    \begin{tabular}{c|l|ccc|cc|c}
    \toprule
    \multicolumn{2}{c}{\multirow{2}{*}{Approaches}} & \multicolumn{3}{c}{Memory} & \multicolumn{2}{c}{Computation} & \multirow{2}{*}{\begin{tabular}[c]{@{}c@{}}Memory \\ Ratio\end{tabular}} \\ \cline{3-7}
    \multicolumn{2}{c}{} & M & O & A & Client & Server &  \\ \midrule
    
     \multirow{2}{*}{\begin{tabular}[c]{@{}c@{}}Activation  \\Reduction\end{tabular}}
     & Down-sampled\textcolor{orange}{~\cite{liu2016ssd}} & - & - & $\downarrow$ & $\downarrow$ & - & * \\
     & Micro-batching\textcolor{orange}{~\cite{NEURIPS2019_Gpipe}} & - & - & $\downarrow$ & - & * & * \\ 
     & Checkpointing\textcolor{orange}{~\cite{Chen_Mem, NEURIPS2019_e555ebe0, MLSYS2020_084b6fbb,sohoni2019low,KARMA}} & - & - & $\downarrow$ & $\uparrow$ & - & * \\ 
     \midrule
    
    \multirow{2}{*}{\begin{tabular}[c]{@{}c@{}}Compr-\\ession\end{tabular}} 
    & Model Pruning\textcolor{orange}{~\cite{arxiv_171001878, PruneTrain}} & $\downarrow$ & $\downarrow$ & - & $\downarrow$ & * &  * \\ 
    & Low Precision\textcolor{orange}{~\cite{MixPrecision}} & $\downarrow$ & $\downarrow$ & $\downarrow$ & $\downarrow$ & $\downarrow$ & $2-4\times$ \\ 
    \midrule
    
    \multirow{2}{*}{\begin{tabular}[c]{@{}c@{}}Model \\ Splitting\end{tabular}} & Layer-wise\textcolor{orange}{~\cite{poirot2019split,thapa2022splitfed}}  & $\downarrow$ & $\downarrow$ & $\downarrow$ & $\downarrow$ & $\uparrow$ & * \\
     & \begin{tabular}[c]{@{}c@{}}\textbf{This work} \end{tabular}
     & $\downarrow$ & $\downarrow$ & $\downarrow$ & - & $\downarrow$ & $\ge 4\times$ \\
    \bottomrule
    \end{tabular}
\end{table*}
\renewcommand{\arraystretch}{1.0}
In summary, the main contributions of this study are threefold:
\begin{itemize}

  \item We propose FedDCT, a novel approach that reduces the memory requirements of training in the FL setting. FedDCT allows lightweight edge devices to overcome resource constraints and participate in the FL task to train large CNN models. To the best of our knowledge, this is the first work that enables the training of large deep learning networks in edge computing devices in FL settings.
  
  \item The proposed FedDCT architecture can divide large CNN networks into sub-networks and collaboratively train them with local clients in parallel. This collaborative training process allows clients from the same cluster to learn representations from each other. The server takes a weighted average of all the ensemble models trained by all the clusters to improve the learning performance of the global model.

  \item We conduct extensive experiments on natural and real-world medical image datasets. FedDCT significantly outperforms a set of the latest state-of-the-art FL methods and sets new state-of-the-art results on these datasets. Our codes are made publicly available at \href{https://github.com/vinuni-vishc/fedDCT}{https://github.com/vinuni-vishc/fedDCT}.
  \item To the best of our knowledge, we are the first to consolidate several advantages of DCML algorithms into a single framework: data privacy, reduced memory demand and asynchronous training, all while achieving superior accuracy compared to other DCML methods.
\end{itemize}

The remainder of our paper is organized as follows. Related works are reviewed in Section~\ref{related-works}. In Section~\ref{preliminaries}, we provide brief preliminaries of an FL setting and an overview formulation for training large CNNs. Section~\ref{proposed-framework} describes the proposed FL framework and the key techniques behind it. 
The experimental results and key properties of FedDCT are presented in Section~\ref{experiments}. Section~\ref{discussion} discusses real-world applications and privacy protection measures regarding FedDCT.  Finally, we conclude the paper and address the key research challenges, limitations of our method, and potential directions of future works in Section~\ref{conclusion}.

\section{Related works}
\label{related-works}
Improving efficiency and effectiveness is extensively researched, but few works address the memory barrier in FL. Existing FL methods such as FedAvg\cite{mcmahan2017communication} and its variant~\cite{asad2020fedopt,wang2020federated, li2020federated2,wang2020tackling} face significant hurdles in training large CNNs on resource-constrained devices. Recent works like FedNAS\cite{he2021fednas} work on large CNNs, but they rely on GPU training to complete the evaluations. Others \cite{bernstein2018signsgd,wangni2018gradient,so2021turbo,lin2017deep} optimize the communication cost without considering edge computational limitations. In Table~\ref{table:overview}, we summarize the techniques for dealing with the memory limitation issue during training. In which each component of memory can be categorized as one of the following~\cite{sohoni2019low}: (i) \textbf{model memory} which consists of the ML model parameters; (ii) \textbf{optimizer memory} which consists of the gradients of the ML model and the hyper-parameters of the optimization methods such as the momentum vectors; and (iii) the \textbf{activation memory} which consists of the intermediate activation, e.g., the output of each layer in the network stored to reuse during the backpropagation phase and its gradients. The activation memory consumes most of the required memory and could far outstrip the model memory, up to $10$s--$100$s times. The optimizer memory is always $2-3$ times larger than the model memory.

\subsection{Training memory reduction techniques}
One of the simplest approaches to reduce the required training memory is to reduce the size of the input samples (down-sampling), which may potentially lead to accuracy degradation\cite{liu2016ssd}. On the other hand, authors in~\cite{NEURIPS2019_Gpipe} proposed reducing the minibatch size by using the micro-batching technique. The mini-batch is divided into smaller micro-batches, which are run in a sequential manner that leads to a reduction in activation size. However, this approach may not work well with other machine learning techniques such as the batch normalization used by Wide-ResNet model~\cite{zagoruyko2016wide}. Alternatively, checkpointing~\cite{Chen_Mem, NEURIPS2019_e555ebe0, MLSYS2020_084b6fbb,sohoni2019low} reduces the activation memory by storing the activations of a subset of layers only and redundantly recomputing the un-stored activations during the backpropagation phases. In the centralized training environment, checkpointing becomes more efficient when combined with the out-of-core method~\cite{KARMA} to avoid too much recomputing by temporarily storing some of the activations in extra memory, e.g., CPU memory. However, such kind of approach is not applicable to the edge devices environment due to the hardware and power consumption restriction. 

\subsection{Compression techniques}
In literature, compression techniques have been well-design for inference on edge-mobile devices ~\cite{huang2019efficient,cai2017deep,sung2015resiliency}. Because most of those techniques often require training with a full model before compressing, it could not help for low-memory training. Alternatively, one can reduce the required training memory by removing the less important connections of the ML networks (model pruning), so that storing fewer parameters with a little or no loss of training accuracy~\cite{arxiv_171001878, PruneTrain}. Another direction aims to train very low-precision, e.g, use a lower number of bits to present a weight value (quantization or low-precision). For example, ~\cite{MixPrecision} demonstrated that training with most storage in half-precision (16 bits) could achieve the same accuracy with 32 and 64-bits numbers. 
\subsection{Memory reduction in Distributed Collaborative Machine Learning}
Split learning (SL)\cite{poirot2019split} attempts to break the computation constraint by splitting the ML model into two portions ${W}^{c}$ and ${W}^{s}$ in a layer-wise manner and offloading the larger portion into the server-side. The communication involves sending activations, called smashed data, at the cut layer of the client-side network to the server, and receiving the gradients of the smashed data from the server side. When doing ML computation on devices with limited resources, assigning only a portion of the network to train on the client side minimizes the processing burden (compared to running a complete network as in FL). In addition, neither a client nor a server can access the other's model. However, the relay-based training in SL idles the clients’ resources because only one client engages with the server at one instance; causing a significant increase in the training time overhead. SplitFed\cite{thapa2022splitfed} (SFL) attempts to solve this by processing the forward propagation and back-propagation on its server-side model with each client’s smashed data separately in parallel, assuming that the server is equipped to handle the training of all clients simultaneously, which is impractical in real-world FL scenarios where millions of device can participate in training\cite{hard2018federated}. FedGKT\cite{he2020group} designs a variant of the alternating minimization approach to train small CNNs on edge nodes and periodically transfer their knowledge by knowledge distillation to a large server-side CNN. As we have reached a state of maturity sufficient to deploy the system in production and solve applied learning problems with millions of real-world devices\cite{bonawitz2019towards}, and we anticipate uses where the number of devices reaches billions, there is a need for a more efficient way to resolve the computation problem in FL without being highly dependent on a central server.

\subsection{Ensemble learning and collaborative learning} 
Ensemble techniques are a ML strategy that integrates many base models to generate a single best prediction model. Some layer splitting algorithms \cite{xie2017aggregated,zhang2020resnest} implicitly employ a ``divide and ensemble" strategy, i.e., they divide a single layer of the model and then fuse the resulting outputs to improve performance. Dropout\cite{srivastava2014dropout} can likewise be regarded as an implicit aggregation of many subnetworks inside a single unified network. Slimmable network\cite{yu2018slimmable} creates many networks with varying widths from a full network and trains them by sharing parameters to obtain adaptive accuracy-efficiency trade-offs at run time. MutualNet\cite{yang2020mutualnet} mutually trains these sub-networks to improve the performance of the entire network. Recent works
\cite{kondratyuk2020ensembling} show that ensembles can outperform single models with both higher accuracy and requiring similar or fewer total FLOPs to compute, even when those individual models' weights and hyperparameters are highly optimized. Collaborative learning methods \cite{zhang2018deep,qiao2018deep,guo2020online} can enhance the performance of individual neural networks by training them with some peers or teachers. Zhao \textit{et al.}\cite{zhao2022toward} shows that increasing the number of networks (ensemble) and collaborative learning can achieve better accuracy-efficiency trade-offs than traditional model scaling methods. Based on this observation, we bring model ensemble and collaborative learning to the FL task.

\section{Preliminaries}
\label{preliminaries}
\begin{figure*}[htbp]
  \centering
  \includegraphics[width=0.9\linewidth]{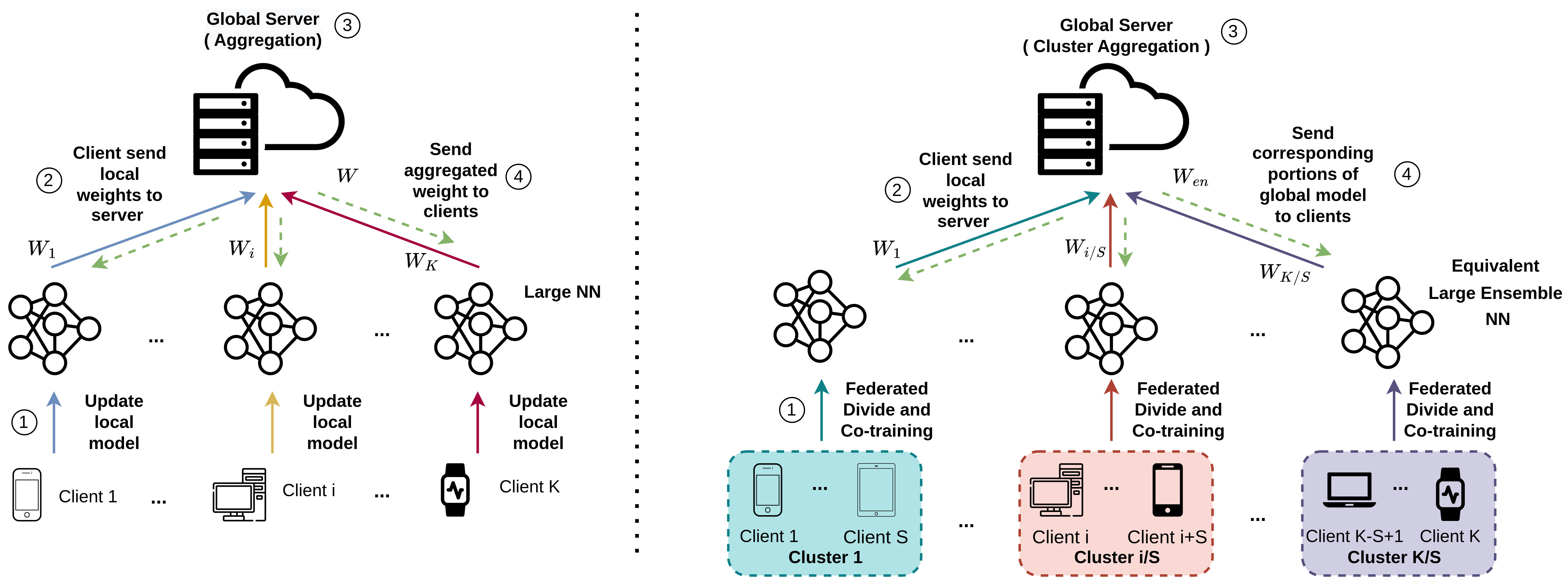}
  \vspace{-0.2cm}
  \caption{Illustration of a typical FL architecture (\textbf{left}) and the proposed FedDCT (\textbf{right}). FedDCT enables a cluster of clients to cooperatively train a big deep learning model by breaking it into an ensemble of several small sub-models and training these sub-models on several devices concurrently while maintaining client data privacy. We refer to this algorithm as Federated Divide and Collaborative Training or FedDCT in short.}
  \label{fig:overview}
\end{figure*} 
We leverage the FL paradigm to collaboratively train large convolutional neural networks on resource-constrained devices (that may not be equipped with GPU accelerators) without transmitting the clients' data to the server.
In the following, we first introduce the traditional FL framework in Section \ref{subsec:baseline}. 
Note that this approach is unsuitable for resource-constrained devices as it requires all clients to train a whole network whose size is usually significant.
To this end, we propose a novel theoretical formulation of FL in Section \ref{subsec:reformulation}, where each device is only responsible for training a portion of the model, and the ensemble learning paradigm is used to build the whole model.

\subsection{Conventional FL framework}
\label{subsec:baseline}
Let us consider the baseline version of the FL algorithm, Federated Averaging (FedAvg) \cite{mcmahan2017communication} as described in Algorithm~\ref{alg:FedAvg}. We denote the set of $K$ mobile devices as $\mathcal{K}$. Each device $k \in \mathcal{K}$ participates in training a shared global model ${W}$ by using its own dataset $\mathcal{D}_{k \in \mathcal{K}} :=\left\{\left({X}_{i}^{(k)}, y_{i}^{(k)}\right)\right\}_{i=1}^{N^{(k)}}$, where ${X}_{i}^{(k)}$ is the $\textit{i}^{th}$ training sample of device $k$, $\textit{y}_{i}^{(k)}$ is the corresponding label of ${X}_{i}^{(k)}$, $y_{i}^{(k)} \in\{1,2, \ldots\}$ (a multi-classification learning task), and $N^{(k)}$ is the cardinality of dataset $\mathcal{D}_{k \in \mathcal{K}}$; $N=\sum_{k \in \mathcal{K}}{N^{(k)}}$.  


At the start of each training round, each device $k$ trains a local model ${W}^{k}$ to minimize a loss function $\mathcal{F}^{(k)}({W})$ as follows
\begin{equation}
\small
{W}^{(k)^*}=\arg \min \mathcal{F}^{(k)}({W}), k \in \mathcal{K}.
\end{equation}
The loss function might vary according to the FL algorithm.
For example, a typical loss function for a multi-classification task can be defined as $\mathcal{F}^{(k)}({W})=\frac{1}{N^{(k)}} \sum_{i=1}^{N^{(k)}} \ell\left({W} ; {X}_{i}^{(k)}, y_{i}^{(k)}\right)$ , where $\ell\left({W} ; {X}_{i}^{(k)}, y_{i}^{(k)}\right)$ quantifies the difference between the expected outcome $y_{i}^{(k)}$ and the result predicted by the model.
After finishing the local training, each device $k$ uploads its computed update ${W}^{k}$ to the central server, which then uses the weighted averaging method to aggregate and calculate a new version of the global model as follows
\begin{equation}
\label{eq:fedavg1}
\small
{W}=\frac{1}{N} \sum_{k=1}^{K}{N^{(k)} {W}^{(k)}}.
\end{equation}
The global model ${W}$ is then sent back to all devices for the next communication round until the global learning is completed. 
\begin{algorithm}[t]
\label{alg:fedavg}
\caption{Federated Averaging (FedAvg). \\
$\mathcal{K}$ is the set of devices, $k$ is a device, $d_k$ is the training data of $k$, and $N^{(k)}$ is the cardinality of $d_k$, $B$ is the local minibatch size, $E$ is the number of local epochs, and $\eta$ is the learning rate.}\label{alg:FedAvg}
\small
\begin{algorithmic}[1]
\Procedure{Server Executes}{}
\State initialize $W_{0} $
\For{each round $t =1,2, \ldots$} 
\State $S_{t} \gets$ (random set of $K$ clients)
\For{each client $k \in S_{t} \text{ in parallel}$} 
\State $W_{t+1}^{(k)} \gets$ ClientUpdate $\left(k, W_{t}\right)$
\EndFor
\State $W_{t+1} \gets \frac{1}{\sum\limits_{k\in \mathcal{K}} N^{(k)}}\sum\limits_{k\in \mathcal{K}} N^{(k)} W_{t+1}^{(k)}$
\EndFor
\EndProcedure
\\
\Procedure{ClientUpdate}{$k,W$} \Comment{Run on client $k$}
\State $\mathcal{B} \leftarrow\left(\right.$ split $d_{k}$ into batches of size $B$ ) 
\For{each local epoch $i$ from $1$ to $E$}
\For{batch $b \in B$} 
\State $W \leftarrow W-\eta \nabla \ell(W ; b)$
\EndFor
\EndFor
\State Return $W$ to server
\EndProcedure
\end{algorithmic}
\end{algorithm}
Consequently, an FL process of a CNN-based global model ${W}$ can be formulated by the following distributed optimization problem
\begin{equation}
\label{eq:fedavg3}
\small
\begin{aligned}
\min _{{W}} \mathcal{F}({W}) \stackrel{\text { def }}{=} \min _{{W}} \sum_{k=1}^{K} \frac{N^{(k)}}{N} \cdot \mathcal{F}^{(k)}({W}),\\ \text { subject to } \mathcal{F}^{(k)}({W})=\frac{1}{N^{(k)}} \sum_{i=1}^{N^{(k)}} \ell\left({W} ; {X}_{i}^{(k)}, y_{i}^{(k)}\right)
.\end{aligned}
\end{equation}

Most federated optimization approaches such as FedAvg\cite{mcmahan2017communication}, FedProx\cite{li2020federated2}, FedNova\cite{wang2020tackling}, and FedMA\cite{wang2020federated} offer to solve objective function Eq. (\ref{eq:fedavg3}) with local SGD (Stochastic Gradient Descent) variants. These methods focus on communication-efficient training and demonstrate their characteristics with experiments on shallow neural networks or linear models. However, the main drawback of these methods is the incapability to train large CNN at lightweight edge devices due to resource constraints such as the lack of GPU accelerators and insufficient memory.

\subsection{A novel FL formulation for training large CNNs}
\label{subsec:reformulation}
To overcome the resource constraints on lightweight edge devices, we propose a novel methodology for the FL optimization problem that can deal with large CNNs training in this section.
As illustrated in Fig.~\ref{fig:overview}, we replace the global CNN model ${W}$ in Eq. (\ref{eq:fedavg3}) with an equivalent ensemble model ${W_{en}}$ of $S$ equal-sized smaller sub-models: ${W}_{en} = \mathcal{E}{\left\{{W}_{1}, {W}_{2}, \ldots, {W}_{S}\right\}}$. 
The principle is to preserve the metrics, i.e., the total number of parameters and FLOPs of all devices' sub-models, to be approximately equivalent to the original undivided model. Consequently, we reformulate a single large global model ${W}$ optimization problem into an optimization problem of the global ensemble model ${W_{en}}$ that requires solving the local objective functions $\left\{\mathcal{F}_{1}, {\mathcal{F}}_{2}, \ldots, {\mathcal{F}}_{s}\right\}$ of the sub-models $\left\{{W}_{1}, {W}_{2}, \ldots, {W}_{S}\right\}$ simultaneously. To enable this, we split set $\mathcal{K}$ of $K$ mobile devices into $K/S$ clusters. We denote the set of clusters of devices as $\mathbb{C}$. Each cluster $C\in \mathbb{C}$ consists of $S$ devices $\left\{{c}_{1}, {c}_{2}, \ldots, {c}_{S}\right\}$. For each communication round, instead of sending the global model ${W}$ to each device $k \in \mathcal{K}$, the server transfers only a portion of the global ensemble model to corresponding devices in each cluster. The devices in each cluster then perform a collaborative training process to obtain an ensemble model. 
These ensemble models are then aggregated by the server to produce the updated global model $W_{en}$.
Consequently, our proposed FL framework can be theoretically formulated as follows
\begin{equation}
\label{eq:fedavg2}
\small
\begin{aligned}
\min _{{W}_{en}} \mathcal{F}({W}_{en})  \stackrel{\text { def }}{=} \min _{{W}_{en}}\frac{1}{\sum\limits_{C \in \mathbb{C}}\left |d_C \right |} \sum\limits_{C \in \mathbb{C}} \left |d_C \right |\cdot \mathcal{F}^{(C)}({W}_{en}),\\
\text { subject to } \mathcal{F}^{(C)}({W}_{en}) = \mathcal{F}^{(C)} \left ( \mathcal{E}\{{W}_{1}, ..., {W}_{s}\} \right ),\end{aligned}
\end{equation}
where $\mathcal{F}^{(C)} \left ( \mathcal{E}\{{W}_{1}, ..., {W}_{s}\} \right )$ is the objective function of cluster $C$.

\textcolor{red}{
\newline
\vspace{2mm} 
}
\textbf{Key Insight} 
The core advantage of the above reformulation is that since the size of a sub-model ${W}_{i}$ ($i = 1, ..., S$) is multiple orders of magnitude smaller than that of the original model ${W}$, the edge training is affordable. Moreover, as discussed in \cite{kondratyuk2020ensembling,zhao2022toward}, instead of using the common practice of tuning a single large model, one can increase the number of networks (ensemble) as a more flexible method for model scaling without sacrificing accuracy. Our experiments prove that employing an ensemble of small models as the global model does not degrade but rather substantially improves FL's performance.\newline

\textbf{Notations and Definitions} In this section, we define the terms used throughout the paper. Model division refers to the channel-wise division of a deep neural network into several smaller neural networks of equal size. Collaborative training is the concurrent training of a large ensemble model by multiple clients, with each client training a portion of the ensemble model. Through an algorithm known as mutual knowledge distillation, these clients can also learn from each other's predictions. A cluster refers to a set of multiple clients who are assigned to collaboratively train one ensemble model. Please refer to table \ref{table:notations} for more details.

\begin{table}[t]
\centering
\small 
\renewcommand{\arraystretch}{1.3} 
\caption{Summary of key notations}
\label{table:notations}
\begin{tabular}{| c | >{\centering\arraybackslash}p{0.7\linewidth} |} 
\hline\hline
\textbf{Notation} & \textbf{Description} \\ [0.3ex] 
\hline
$\mathcal{{K}}$ & Set of $K$ devices chosen for current training round\\ \hline
$\mathbb{C}$ & The set of all $\frac{K}{S}$ device clusters. Each cluster $C \in \mathbb{C}$ consists of $S$ devices $c_1, ..., c_S$ \\ \hline
$W_{en}$ & Global ensemble model of $S$ equal-sized sub-models $W_{1}, ..., W_{S}$ \\ \hline
${W}_{i}^{m}$ & Lower sub-model, consists of the few first layers of ${W}_{i}$\\ \hline
${W}_{i}^{p}$ & Upper sub-model, consists of the remaining layers of the ${W}_{i}$\\ \hline
$\boldsymbol{X}_{i}^{(k)}$ & The $\textit{i}$th training sample of device $c_k$   \\ \hline
$\textit{y}_{i}^{(k)}$ &  The corresponding label of sample $\boldsymbol{X}_{i}^{(k)}$   \\ \hline
$\mathcal{L}_{c e}\left(p_{i}, y_{k}\right)$ &  Cross-entropy loss of device $c_i$ with $c_k$ as main client \\ \hline
$\mathcal{L}_{\mathrm{cot}}$ & Collaborative training loss of clients in a cluster  \\ \hline
$\mathcal{F}^{(C)}({W}_{en})$ & Final objective function with client $c_k$ as the main client  \\ \hline
$W_{C}$ & Updated ensemble model of cluster $C$ \\ \hline
$N^{(C)}$ & The total number of samples accross all devices in cluster $C$  \\ \hline
\hline
\end{tabular}
\end{table}

\begin{figure*}[htbp]
  \centering
  \includegraphics[width=0.9\linewidth]{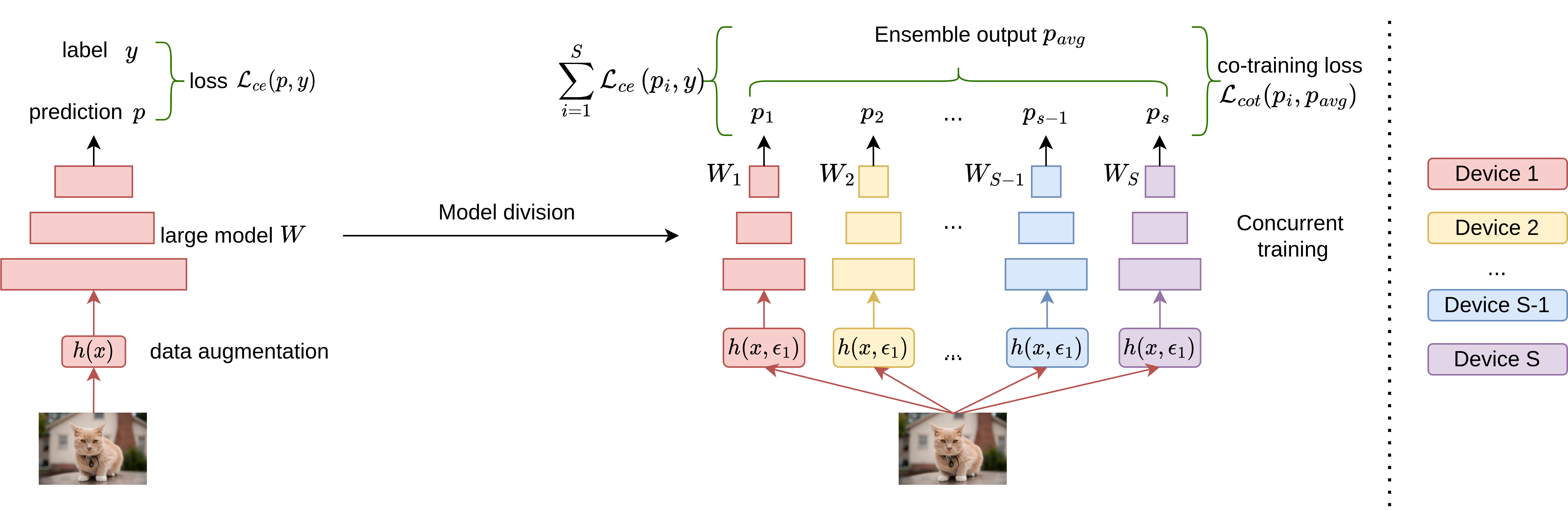}
  \vspace{-0.2cm}
  \caption{Model division and concurrent training on multiple devices. This division and collaborative training approach will not work in an FL environment since devices cannot directly exchange data. Accordingly, we propose a novel Federated Divide and Collaborative Training paradigm. }
  \label{fig:modeldivision}
\end{figure*} 
\section{The Proposed Framework}
\label{proposed-framework}
Based on the above reformulation, we propose FedDCT, a novel method for distributed training of large-scale neural networks on lightweight edge devices under the FL settings. First, we discuss the model division and collaborative training techniques in Sections \ref{subsec:model division} and \ref{subsec:co-training details}, respectively. We outline the FedDCT workflow based on these methods in Section \ref{subsec:overview}, and we then go into depth about the federated collaborative training and cluster aggregation procedure in Sections \ref{subsec:co-training} and \ref{subsec:clusteragg}, respectively. Section \ref{subsec:inference} will discuss the inference phase of FedDCT.

\subsection{Model division}
\label{subsec:model division}
Given one large network, we first (channel-wise) divide it according to its width, or more specifically, the parameters or FLOPs of the network. For instance, if we want to divide a network into four small networks, the number of parameters of one small network will become one-fourth of the original. Let ${W}$ be the original model. As demonstrated in Fig.~\ref{fig:modeldivision}, we want to divide ${W}$ into an ensemble of $S$ small sub-models $\left\{{W}_{1}, \ldots, {W}_{S}\right\}$. Our goal is to keep the metrics: the total number of parameters and FLOPs roughly unaltered before and after division.\newline

\textbf{Parameters.} 
We view ${W}$ as a stack of convolutional layers. Following Pytorch's definition\cite{Conv2dpytorch}, consider a convolutional layer with $M\times M$ as the kernel size, $C_{in}$ and  $C_{out}$ as the number of channels of input and output feature maps, respectively, and $d$ as the number of groups (i.e., every $\frac{C_{in}}{d}$ input channels are convolved with its own sets of filters, of size $\frac{C_{out}}{d}$). In the case of depthwise convolution\cite{liu2016ssd}, $d = C_{in}$. Then, the number of parameters and FLOPs of that convolutional layer can be derived as follows
\begin{equation}
\small
\text { Params: } M^{2} \times \frac{C_{i n}}{d} \times \frac{C_{\text {out }}}{d} \times d,
\end{equation}
\begin{equation}
\small
\text { FLOPs : }\left(2 \times M^{2} \times \frac{C_{i n}}{d}-1\right) \times H \times W \times C_{\text {out },}
\end{equation}
where $H \times W$ is the size of the output feature map. Bias is omitted here for brevity. Generally, $C_{out} = t_1 \times C_{in}$, where $t_1$ is a constant. Therefore, the number of parameters is equivalent to
\begin{equation}
\small
\text { Params: } M^{2} \times t_{1} \times\left({C_{in}}\right)^{2} \times \frac{1}{d}.
\end{equation}

Therefore, in order to divide a convolutional layer by a factor of $S$, we just need to divide $C_{in}$ by $\sqrt{S}$ as follows 
\begin{equation}
\small
\frac{M^{2} \times t_{1} \times\left({C_{in}}\right)^{2}}{S} \times \frac{1}{d}=M^{2} \times t_{1} \times\left(\frac{C_{i n}}{\sqrt{S}}\right)^{2} \times \frac{1}{d}.
\end{equation}
For instance, if we divide a bottleneck layer in ResNet
by a factor of 4, it becomes 4 small blocks as 

\begin{equation}
\label{eq:divide}
\small
\left[\begin{array}{cc}1 \times 1, & 64 \\ 3 \times 3, & 64 \\ 1 \times 1, & 256\end{array}\right] \xrightarrow[\text{to}]{\text{equals}} 4 \times \left[\begin{array}{cc}1 \times 1, & 32 \\ 3 \times 3, & 32 \\ 1 \times 1, & 128\end{array}\right]
\end{equation}

Compared to the original block, each small block in Eq.~(\ref{eq:divide}) has only a quarter of the parameters and FLOPs. More details about model division is given in the \ref{appendix}\textcolor{blue}{ppendix}.

\textbf{Regularization} 
Since the model capacity degrades after partitioning, the regularization components in networks should be adjusted correspondingly. To this end, we change the scale of dropping regularization linearly based on the assumption that model capacity is linearly dependent on network width. Specifically, we divide the dropout's dropping probabilities\cite{srivastava2014dropout} and stochastic depth\cite{huang2016deep} by $\sqrt{S}$ in the experiments. Due to its unclear intrinsic mechanism, the weight decay value is kept unchanged. It is worth noting that the model partition described above is constrained by a number of factors, such as the number of parameters in the original model must be divisible by $S$, and $\sqrt{S}$ must be an integer. In practice, however, these restrictions are not always satisfied. For those cases, we simply round the values $\frac{C_{in}}{\sqrt{S}}$ to the nearest integer numbers.

\textbf{Concurrent training}: For a large ensemble network to be trained concurrently on multiple devices, we follow the strategy as shown in Fig.~\ref{fig:modeldivision}. In Sections \ref{subsec:overview} and \ref{subsec:co-training}, we will discuss on how to implement this training strategy in a data-privacy aware manner in FL setting.
\subsection{Collaborative Training strategy}
\label{subsec:co-training details}

After dividing, small networks are trained with different views of the same data to increase diversity. This allows each sub-network to learn from each other to further boost individual performance. In the following, we describe in detail our training strategy, including data augmentation and model initialization for enhancing the ensemble diversity; and mutual learning loss to mutually exploit knowledge between the sub-models.

\textbf{Different initialization and data views.}
The difference in initialization of the sub-models is one of the factors affecting the effectiveness of an ensemble learning scheme. 
In general, an increase in variety helps to capture different types of information, hence enhancing the ensemble model's efficiency. Different initialization across sub-models is achieved in our proposed model using two techniques: Weight initialization and data augmentation.
We first initialize the sub-models with different weights\cite{li2018research}. Moreover, during training, each sub-model is also provided with a different abstract representation of the same input image. In particular, for each sub-network ${W_i}$ in Fig.~\ref{fig:modeldivision}, we use a random data transformer with $\epsilon_i$ as the random seed to generate a separate view $h(x, \epsilon_i)$ of the input image $x$. In practice, different data views are generated by randomness in the data augmentation procedure. Besides the commonly used random resize, crop, and flip, we further introduce random erasing\cite{zhong2020random} and random augmentation~\cite{cubuk2020randaugment} in our algorithm design. We can guarantee that $h(x, \epsilon_1), h(x, \epsilon_2), \ldots, h(x, \epsilon_ s)$ will produce different views of $x$ for corresponding networks in most cases by applying these random data augmentation operations multiple times. 
\begin{algorithm}
\caption{Federated Divide and Cotraining (FedDCT). \\
$\mathcal{K}$ is the set of all devices, ${W}_{C}$ is the model of cluster $C$, $N^{(C)}$ is the number of data samples owned by devices in cluster $C$, $L$ is the cutting layer, $E$ is the number of local epochs, and $\eta$ is the learning rate.}\label{alg:FedDCT}
\label{algo2}
\begin{algorithmic}[1]
\small
\Procedure{Server Executes}{}
\State initialize ${W}_{en_{0}}^{m} = \mathcal{E}{\left\{{W}_{1_{0}}^{m}, {W}_{2_{0}}^{m}, \ldots, {W}_{S_{0}}^{m}\right\}}$ 
\State initialize ${W}_{en_{0}}^{p} = \mathcal{E}{\left\{{W}_{1_{0}}^{p}, {W}_{2_{0}}^{p}, \ldots, {W}_{S_{0}}^{p}\right\}}$
\For{each round $t =0,1, \ldots$} 
\State $K_{t} \gets$ (random set of $K$ clients from $\mathcal{K}$)
\State $\mathbb{C} \gets$ (partition the $K_{t}$ into to $\frac{K}{S}$ clusters)
\State Sends ${W}_{en_{t}}^{m}$ to first device in each cluster
\State Sends ${W}_{i_{t}}^{p}$ to $i^{th}$ device in each cluster
\For{each cluster $C \in \mathbb{C} \text{ in parallel}$} 
\State ${W}^{m}_{C_{t+1}},W^{p}_{C_{t+1}} \gets$ FedCoTraining $\left(C, {W}_{{en}_{t}}^{m},W_{{en}_{t}}^{p}\right)$\Comment{Update lower ensemble model and upper ensemble model}
\State ${W}_{C_{t+1}} \gets$ Merge $({W}^{m}_{C_{t+1}},{W}^{p}_{C_{t+1}})$
\EndFor
\State ${W}_{en_{t+1}}= \sum\limits_{C \in \mathbb{C}} \frac{N^{(C)}}{N} {{W}}_{C_{t+1}}$
\EndFor
\EndProcedure
\\
\Procedure{FedCoTraining}{$C$,${W}^{m}$,${W}^{p}$} \Comment{Run on cluster $C$ of $S$ devices}
\For{each local epoch from $1$ to $E$}
\For{each device $c_{i} \in C$ ($i{ }=1,2, \ldots, S$}) 
\State $a({W}^{m}), y \gets $ MainDeviceForward $\left(c_{i},{W}^{m}\right)$
\State ${W}^{p}, d\mathbf{A}({W}^{p}) \gets$ ProxyDevicesUpdate $\left(a({W}^{m}),y,C,{W}^{p}\right)$
\State $ {W}^{m} \gets $MainDeviceBackProp($d \mathbf{A}({W}^{p})$)
\State Sends ${W}^{m}$ to device $c_{i+1}$
\EndFor
\EndFor
\State Last device $c_{S}$ sends ${W}^{m}$ to server
\State Each device $c_{i}$ sends ${W}^{p}_{i}$ to server
\EndProcedure
\\
\Procedure{MainDeviceForward}{$c_{i},{W}^{m}$} \Comment{Run on device $c_{i}$}
\For{each data $X_{j} \in c_{i}$}
\State $\left \{ h(X_{j},\epsilon_{1}),h(X_{j},\epsilon_{2}),\ldots,h(X_{j},\epsilon_{S}) \right \} \gets$RandAug$(X_{j})$
\For{each view $h(X_{j},\epsilon_{k}), k =1,2, \ldots, S$}

\State $a({W}^{m}_{k}) \gets $Forward propagation with $h(X_{j},\epsilon_{k})$ up to layer $L$ in ${W}^{m}_k$\Comment{produce $S$ abstract representations}

\State Send $a({W}^{m}_{k})$ and $y_{j}$ ( true label of $X_{j}$) to device $c_{k}$
\EndFor
\EndFor
\EndProcedure
\\
\Procedure{ProxyDevicesUpdate}{$\left(a({W}^{m}),y,C,{W}^{p}\right)$}
\For{each device $c_{k} \in C$, $k =1,2, \ldots, S $ \text{in parallel}} 
\State $p_{k} \gets $ Forward propagation with $a({W}^{m}_{k})$ on ${W}^{p}_{k}$ , computes prediction
\State $\mathcal{L}_{ce}(p_{k}, y) \gets $ Loss of ${W}^{p}_{k}$ concerning $\left(a({W}^{m}_{k}),y \right)$
\State Client $c_{k}$ sends  $p_{k}$ to the server and gets $\mathcal{L}_{cot}$
\State Backpropagation calculates $\nabla \ell_{k} \left ({{W}}_{k}^{p},a({W}_{k}^{p})\right)$
\State $d \mathbf{A}({W}_{k}^{p}) \gets \nabla \ell_{k} \left ({{W}}_{k}^{p},a({{W}}_{k}^{p})\right)$
\State Send $d \mathbf{A}({W}_{k}^{p})$ to Main Device for MainDeviceBackprop($d \mathbf{A}({W})$)
\State ${W}^{p}_{k} \leftarrow {W}^{p}_{k}-\eta \nabla \ell_{k} \left ({{W}}_{k}^{p},a({{W}}_{k}^{p})\right)$
\EndFor
\EndProcedure
\\
\Procedure{MainDeviceBackprop}{$d \mathbf{A}({W}) $}
\For{ each Gradient $d \mathbf{A}({W}_{g})$ , $g=1,2, \ldots, S $ }
\State Backpropagate, calculates gradients $\nabla \ell_{g} ({{W}}_{g}^{m})$ with $d \mathbf{A}({W}_{g})$ 
\State ${W}^{m}_{g} \leftarrow {W}^{m}_{g}-\eta \nabla \ell_{g} ({{W}}_{g}^{m})$
\EndFor
\EndProcedure
\end{algorithmic}
\end{algorithm}

\textbf{Mutual knowledge distillation}. We can further boost the ensemble performance by allowing sub-models to learn from each other, thereby avoiding poor performance caused by overly de-correlated sub-networks\cite{qiao2018deep}. Following the co-training assumptions\cite{blum1998combining}, sub-networks are expected to agree on their predictions of $x$ although they see different views of $x$. In particular, we adopt a natural measure of similarity, the Jensen-Shannon divergence\cite{menendez1997jensen} between predicted logits of sub-networks, as the co-training objective
\begin{equation}
\small
    \mathcal{L}_{\mathrm{cot}}\left({p}_{1}, {p}_{2}, \ldots, {p}_{S}\right) = H\left(\frac{1}{S} \sum_{k=1}^{S} {p}_{k}\right)-\frac{1}{S} \sum_{k=1}^{S} H\left({p}_{k}\right),
\end{equation}
\newline
where ${p}_{k}$ depicts the distribution of sub-model $M_{k}$'s predicted logits for input $h(x,\epsilon_{k})$, and $H({p})=\mathbb{E}[-\log ({p})]$ represents the Shannon entropy\cite{shannon1948mathematical} of a distribution ${p}$. The global objective function is the sum of the classification losses $\mathcal{L}_{ce}$ of small networks and the co-training loss. During inference, we average the outputs before
softmax layers as the final ensemble output.
\subsection{FedDCT workflow}
\label{subsec:overview}
To elaborate, we illustrate the overall workflow in Fig.~\ref{fig:overview} as well as summarize the procedures of FedDCT in Algorithm \ref{algo2}. First, the global model is channel-wise divided into an ensemble of $S$ sub-models regarding its parameters and regularization components. The model division is explained in depth in Section \ref{subsec:model division}. Additionally, the model division procedure is modified to comply with FL's privacy standards. At the beginning of each communication round, $K$ clients are randomly selected to participate in the training procedure. These $K$ devices are grouped into $K/S$ clusters of $S$ devices. To initialize the network, the server sends appropriate portions of the global model to each participating client. Then, each cluster updates the ensemble model in parallel using only local data from its clients. Specifically, clients from each cluster collaborate to simultaneously train the large ensemble model \textbf{without} exchanging data in a process we refer to as ``Federated Collaborative Training". Using FedAvg, the server aggregates updated ensemble models from all clusters to generate a new global ensemble model. During testing, each client will be responsible for running the whole ensemble model (by running inference on each sub-model sequentially) or send these intermediate outputs (without sharing data) to the
server to complete the inference process through a network connection. We take the mean of outputs before softmax layers of all sub-models as the final output of the model.
\begin{figure*}[]
\centering
\includegraphics[width=0.8\linewidth]{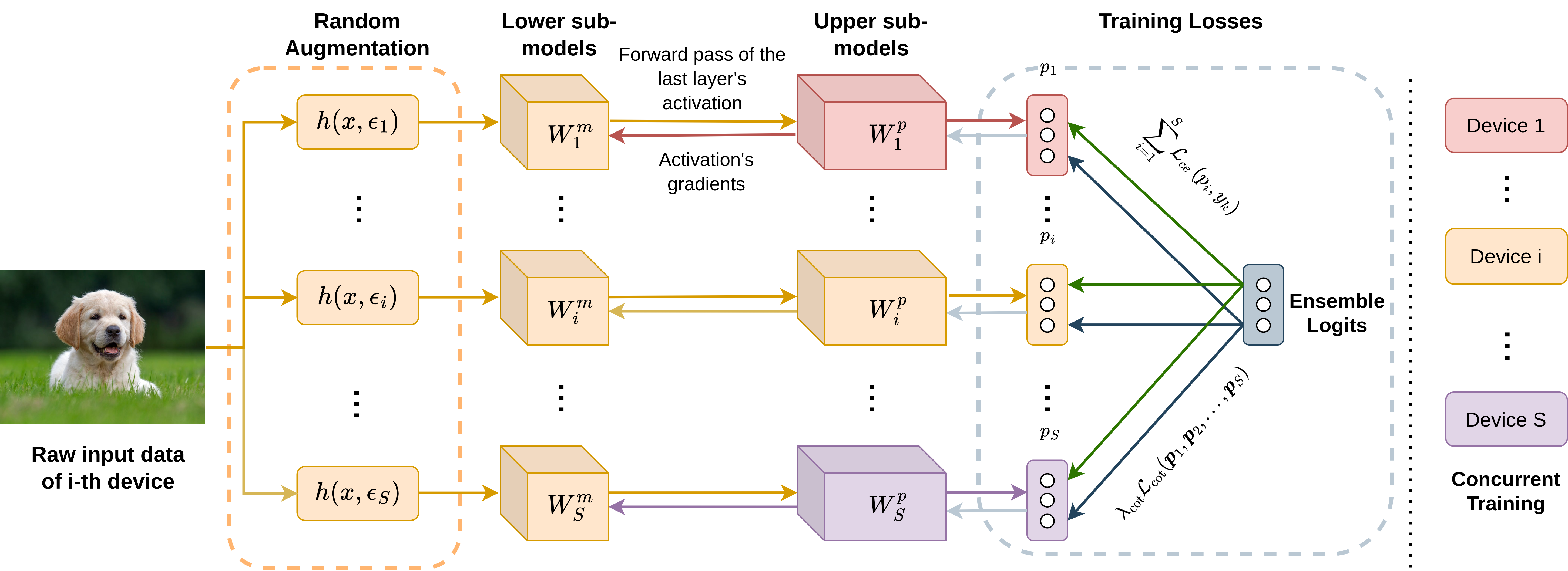}
\caption{
\normalsize Federated Collaborative Training of an ensemble model across $S$ clients in a cluster of FedDCT. Let $x$ be a training data of main device $c_i$, then $c_i$ first creates $S$ augmented data and run forward propagation through  $\left\{{W}_{1}^{m}, {W}_{2}^{m}, \ldots, {W}_{s}^{m}\right\}$ to obtain $S$ abstract representations. These representations are then sent to upper sub-models of other clients for collaborative training.}
\label{fig:divideandcotrain}
\end{figure*}
\subsection{Federated Divide and Collaborative Training}
\label{subsec:co-training}
\subsubsection{\textbf{Model division in FL setting}}
\label{subsubsec:feddiv}
Let us refer to the set of $S$ devices of a cluster $C$ as $c_1, c_2, ..., c_S$. As stated previously, the original model ${W}$ is divided into an ensemble of $S$ sub-models ${W}_{en} = \mathcal{E}{\left\{{W}_{1}, \ldots, {W}_{S}\right\}}$, that will be concurrently trained by $S$ devices. The naive approach to collaborative training would be to send the training data directly from one client to the others in a cluster in order to update the sub-models simultaneously (see Fig.~\ref{fig:modeldivision}). However, this defies FL's emphasis on privacy, as data is explicitly shared among clients. To reap the benefits of collaborative training while ensuring client's data privacy, we propose a novel federated collaborative training paradigm. This process is illustrated in Fig.~\ref{fig:divideandcotrain}.

Initially, we define the \emph{cut layer} as an intermediate layer of ${W}_i$ ($i=1, ..., S$), which layer-wise split ${W}_i$ into two portions:
\begin{itemize}
    \item Lower sub-model ${W}_{i}^{m}$, consists of only a few first layers of ${W}_{i}$ which are responsible for extracting the abstract representation of the input.
    \item Upper sub-model ${W}_{i}^{p}$, includes the remaining layers of the ${W}_{i}$ which are responsible for making the prediction.
\end{itemize}

At the beginning of each communication round, we choose the first client (device $c_1$) in each cluster $C \in \mathbb{C}$ as the main client of that cluster. The remaining clients in the cluster are now considered proxy clients. The server sends ${W}^{m}_{en} = \mathcal{E}{\left\{{W}^{m}_{1}, \ldots, {W}^{m}_{S}\right\}}$ to the main client in each cluster and ${W}_{i}^{p}, (i=1, ..., S)$ to the $i^{th}$ client in each cluster.\newline

\subsubsection{\textbf{Federated Collaborative Training}}
\label{subsubsec:fedcot}
Given a set of input-output pairs $(X_{1}, y_{1})$, where $X_{1}$ is the data sample of main device $c_{1}$, and $\textit{y}_{1}$ is the corresponding label of $X_{1}$. Instead of sending $X_{1}$ directly to each device in $ \left\{{c}_{1}, {c}_{2}, \ldots, {c}_{s}\right\} $ to complete the training process of $\left\{{W}_{1}, {W}_{2}, \ldots, {W}_{s}\right\}$, $c_1$ will generate $S$ augmented versions of ${X}_{1}$ and execute forward propagation on $\left\{{W}_{1}^{m}, {W}_{2}^{m}, \ldots, {W}_{s}^{m}\right\}$ to produce $S$ unique abstract representations of the same input image. In other words, the main client performs forward propagation until the cut layer of $S$ lower sub-models utilizing $S$ different view of the same input image. The activations (smashed data) at $S$ cut layers are then sent to $S-1$ proxy clients in the cluster, with one abstract representation remaining on the main client.

Each device ${{c}_i} \in \left\{{c}_{1}, {c}_{2}, \ldots, {c}_{s}\right\} $ then completes the forward propagation on corresponding upper sub-model ${W}_{i}^{p}$ with smashed data $a({W}_{i}^{m})$ separately in parallel with other clients, and sends its prediction result $p_{i}$ to the server. This completes a round of forward propagation without sharing raw data. The server then gathers the predictions $\left\{p_{1}, p_{2}, \ldots, p_{S}\right\}$ from all devices, and calculate the cross entropy classification loss of the entire ensemble model as $\sum_{i=1}^{S} \mathcal{L}_{ce}(p_{i}, y_{1})$. Furthermore, to improve the generalization performance, we add a regularization term that allows the devices to mutually learn from each other. This regularization term is designed as the Jensen-Shannon (JS) divergence \cite{menendez1997jensen} among predicted probabilities. The final objective function with $c_1$ as the main client now becomes
\begin{equation}
\label{eq:objectivefunc}
\small
\mathcal{F}^{(C)}({W}_{en})=\sum_{i=1}^{S} \mathcal{L}_{c e}\left(p_{i}, y_{1}\right) + \lambda_{\cot } \mathcal{L}_{\mathrm{cot}}\left({p}_{1}, {p}_{2}, \ldots, {p}_{S}\right).
\end{equation}

where ${p}_{i}$ ($i=1, ..., S$) is a vector representing the output of ${W}_{i}^{p}$ on the data sample of main device $c_{1}$, $\mathcal{L}_{cot}$ is the collaborative training loss (the details of $\mathcal{L}_{cot}$ is given in Section \ref{subsec:co-training details}), $\lambda_{cot}$ is the weight factor of $\mathcal{L}_{cot}$ and $\mathcal{L}_{ce}$ is the cross entropy classification loss. $\mathcal{F}^{(C)}({W}_{en})$ is then sent back to all devices. Each device ${c}_{i}$ then performs backpropagation on its upper sub-model ${W}_{i}^{p}$. In other words, gradients are back-propagated from the last layer to the cut layer of each sub-model at the corresponding client. The gradients at the cut layers $\nabla \ell_{i}({W}_{i}^{p},a({W}_{i}^{p}))$  are sent back to the main client, which then completes the rest of backpropagation on $\left\{{W}_{1}^{m}, {W}_{2}^{m}, \ldots, {W}_{s}^{m}\right\}$. This completes a round of forward and backward passes without looking at each other's raw data. This procedure is repeated for a number of epochs. The main client then transfer the updated ${W}^{m}_{en} = \mathcal{E}{\left\{{W}^{m}_{1}, \ldots, {W}^{m}_{S}\right\}}$ to the next client in the cluster ($c_2$ in the case of $c_1$). This client will become the new main client, and training will be repeated with a new main client. This procedure proceeds until each client in the cluster finish training as the main client once. The overall final objective function with $k^{th}$ client $c_k$ as the main client is
\begin{equation}
\label{eq:objectivefuncgeneral}
\small
\mathcal{F}^{(C)}({W}_{en})=\sum_{i=1}^{S} \mathcal{L}_{c e}\left(p_{i}, y_{k}\right) + \lambda_{\cot } \mathcal{L}_{\mathrm{cot}}\left({p}_{1}, {p}_{2}, \ldots, {p}_{S}\right).
\end{equation}

\subsection{Cluster Aggregation}
\label{subsec:clusteragg}
After the training round, the last main client $c_S$ in each cluster $C \in \mathbb{C}$ then sends an updated version ${W}^{m}_{C}$ of global lower ensemble model ${W}^{m}_{en}$ to the server, and each device $c_{i}$ in cluster $C$ sends the corresponding updated upper sub-model ${W}_{i}^{p}$ to the server. The server will then ensemble the received weights from the clients to form the corresponding ensemble model of each cluster. Specifically, for each cluster $C$, corresponding upper sub-models ${W}_{1}^{p}, ..., {W}_{S}^{p}$ are ensembled to generate an upper ensemble model ${W}_{C}^{p}=\mathcal{E}\left\{{W}_{1}^{p}, {W}_{2}^{p}, \ldots, {W}_{s}^{p}\right\} $ of that cluster. The server then merges ${{W}_{C}^{m}}$ and ${{W}_{C}^{p}}$ to create an ensemble model ${{W}_{C}}$ for each cluster. The server then aggregates the updated ensemble models of all clusters to produce a new global ensemble model: 

\begin{equation}
\small
\begin{aligned}
{{W_{en}}}=\frac{1}{N} \sum\limits_{C \in \mathbb{C}} N^{(C)} {W}_{C},
\end{aligned}
\end{equation}
where $N^{(C)}$ is the cardinality of cluster $C$. This concludes a communication round in FedDCT. Therefore, a communication round in FedDCT begins when the selected clients are divided into clusters and receive sub-models, and ends when the server aggregates the ensemble models from all clusters to generate a new global model.
\subsection{Inference phase}
\label{subsec:inference}
Unlike other SOTA works \cite{thapa2022splitfed, he2020group, poirot2019split} where due to memory constraints, inference (on low-resource clients) can only be done online with the help of a central server, FedDCT supports both online and offline on-client inference. In the inference phase, the client can execute forward propagation on $\left\{{W}_{1}^{m}, {W}_{2}^{m}, \ldots, {W}_{s}^{m}\right\}$ and send these intermediate outputs (without sharing data) to the server to complete the inference process through a network connection. If local inference is preferred, the client can download the upper sub-models $\left\{{W}_{1}^{p}, {W}_{2}^{p}, \ldots, {W}_{s}^{p}\right\}$ locally and run inference on each sub-model sequentially to avoid memory constraints. 
\section{Experiments and Results}
\label{experiments} 
In this section,  we extensively evaluate the performance of FedDCT on image classification tasks. The proposed framework achieves higher performance as well as more stable convergence compared with state-of-the-art FL methods~\cite{li2020federated2,mcmahan2017communication,thapa2022splitfed,he2020group} on both standard natural datasets~\cite{cifar} and real-world medical image datasets\cite{tschandl2018ham10000}. In the following, we firstly introduce the four image classification datasets used in our experiments and describe how the data is split over the devices\footnote{In the following, we use the terms \emph{``device"} and \emph{``client"}, interchangeably}. We then describe the setup for the experiments and implementation details of the FedDCT on each dataset. After that, we report and compare our proposed FedDCT with current state-of-the-art FL methods using Top-1 accuracy. We chose Top-1 Accuracy as our metric due to its simplicity and widespread usage in SOTA federated learning techniques\cite{thapa2022splitfed,mcmahan2017communication,li2020federated2,he2020group}. Finally, we conduct ablation studies to highlight some important properties of the proposed framework.
\begin{figure*}
	\centering
	\begin{minipage}[!t]{0.41\textwidth}
	    \centering
         \includegraphics[height=3cm]{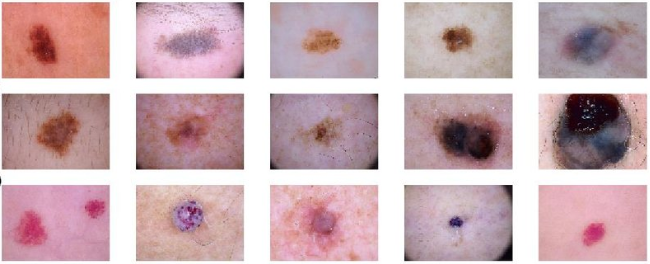}
         \caption{Representative examples of skin cancer images from HAM10000 dataset~\cite{tschandl2018ham10000}.}
         \label{fig:pill-ham}
 	\end{minipage}
	\hspace{20pt}
	\begin{minipage}[!t]{0.45\textwidth}
	    \centering
        \includegraphics[height=2.5cm]{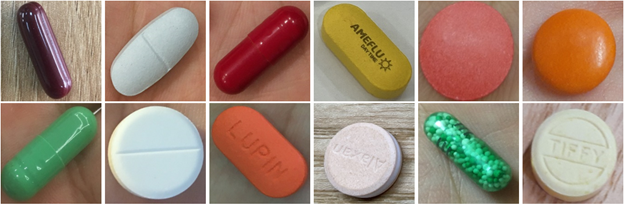}
        \caption{Representative examples of pill images from the VAIPE  dataset.}
        \label{fig:pill-image}
 	\end{minipage}
 	\vspace{-0.2cm}
\end{figure*}

\subsection{Datasets and experimental settings} 
We evaluate the effectiveness of FedDCT on standard image classification datasets including CIFAR-10~\cite{cifar}, CIFAR-100~\cite{cifar}, and two real-world medical imaging datasets that are HAM10000~\cite{tschandl2018ham10000} and VAIPE. For each dataset, we use the official training and validation splits as proposed in the original papers~\cite{cifar,tschandl2018ham10000} for all our experiments.

\textbf{CIFAR-10}~\cite{cifar}. The CIFAR-10 dataset consists of 60,000 color images in 10 classes, with 6,000 images per class. The dataset provides 50,000 training images and 10,000 test images with a size of 32$\times$32 pixels. On CIFAR-10, we consider this experiment as a balanced and IID setting. In particular, all training examples are divided into 20 clients, each containing 2,5000 training examples. CIFAR-10 is available at~\href{https://www.cs.toronto.edu/~kriz/cifar.html}{https://www.cs.toronto.edu/~kriz/cifar.html}.

\textbf{CIFAR-100}~\cite{cifar}. To validate FedDCT further on a dataset with a larger number of classes, we conduct experiments on the CIFAR-100. The CIFAR-100 dataset is similar to CIFAR-10, except that it has $100$ categories each containing $600$ images.
The number of images for training and testing per class is $500$ and $100$, respectively. The training images are equally partitioned into $20$ clients, each containing $2,500$ samples. This setting also is a balanced IID setting. CIFAR-100 dataset is available at~\href{https://www.cs.toronto.edu/~kriz/cifar.html}{https://www.cs.toronto.edu/~kriz/cifar.html}. 

\textbf{HAM10000}~\cite{tschandl2018ham10000}. HAM10000 (``Human Against Machine with 10,000 training images'') is a large-scale database of multi-source dermatoscopic images of common pigmented skin lesions collected from different populations, acquired and stored by different modalities. The main task of this dataset is to identify critical diagnostic categories in the realm of pigmented lesions. In total, the HAM10000 dataset contains $10,015$ published dermatoscopic images; each has a dimension of $600 \times 450$ pixels (see examples in Fig.~\ref{fig:pill-ham}). As a real-world medical imaging dataset, HAM10000 is a high-class imbalanced dataset, with the minority class being rare diseases. In our experiments, we randomly use $80\%$ of the images for training and the rest of $20\%$ for evaluation. Similar to CIFAR-10 and CIFAR-100, we partition the training dataset into 20 clients, each containing about $500$ images. The HAM10000 Dataset is available for download at~\href{https://dataverse.harvard.edu/dataset.xhtml?persistentId=doi:10.7910/DVN/DBW86T}{https://www.kaggle.com/datasets/kmader/skin-cancer-mnist-ham10000}.

\textbf{Pill Image Identification Dataset (VAIPE)}. To further evaluate the robustness of the proposed method in a real-world setting, we have collected and annotated a large-scale pill image identification dataset. The dataset, which we call VAIPE, is a real-world pill image dataset consisting of $98$ pill categories. The main task of the VAIPE dataset is to classify pill images correctly into classes. In our experiments, $8,161$ images are used as training examples, and $1,619$ images are used for evaluation. 
The VAIPE dataset contains pill images taken in real-world scenarios under unconstrained environments, including different lighting, backgrounds, and devices. 
 Fig.~\ref{fig:pill-image} shows several representative examples from the VAIPE dataset. For VAIPE, we also equally divide the training samples to 20 clients.
 Experiments on such a dataset allow us to investigate the robustness of FedDCT in real-world scenarios.
 To make this work reproducible and encourage new advances, we make all images and annotations of the VAIPE dataset publicly available as a part of a bigger dataset that we released on our project website at \href{https://smarthealth.vinuni.edu.vn/resources/}{https://smarthealth.vinuni.edu.vn/resources/}.

\subsection{Implementation details }
\begin{figure*}[]
\centering
    \includegraphics[width=\linewidth]{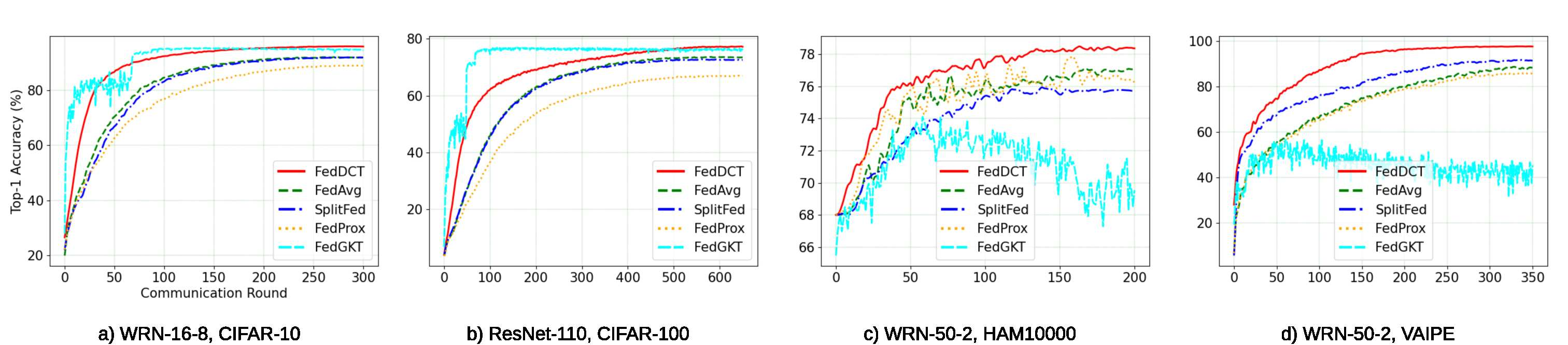} 
    \vspace{-0.6cm}
\caption{Top-1 accuracy (\%) of FedDCT compared to state-of-the-art FL methods on the test sets of CIFAR-10, CIFAR-100, HAM10000, and VAIPE datasets.} 
\label{fig:1}
\end{figure*}
\label{implementation_details}
This section presents our implementation details for experiments across all four datasets. WRN-16-8~\cite{zagoruyko2016wide}, ResNet-110~\cite{he2016deep}, and WRN-50-2~\cite{zagoruyko2016wide} are used as baseline networks. 
To implement these networks, we follow  the original implementations described in ~\cite{he2016deep}~\cite{zagoruyko2016wide}. For all experiments, the networks first initialize the weights using the Kaiming initialization technique~\cite{he2015delving}. We then train the networks using Nesterov accelerated SGD optimizer with a momentum of $0.9$. We apply the warm-up and cosine learning rate decay policy~\cite{loshchilov2016sgdr}. To verify the effectiveness of the proposed method, we compare the performance of FedDCT (S = 4) with state-of-the-art FL approaches, including FedProx~\cite{li2020federated2}, FedAvg~\cite{mcmahan2017communication}, SplitFed~\cite{thapa2022splitfed} and FedGKT\cite{he2020group}. ``S" here refers to the split factor of the global model, and correspondingly the number of clients in a cluster. To this end, we reproduce these approaches on four datasets. For a fair comparison, we apply the same training methodologies and hyper-parameter settings as described in the original papers~\cite{li2020federated2,mcmahan2017communication,thapa2022splitfed,he2020group}. Additionally, SplitFed, FedDCT and FedGKT uses the same cut-off layer for the client-side model in all experiments. We also compare the proposed framework's performance with the centralized training approach. In all experiments, the number of local epochs in each round is set to 1.\newline 
\textbf{Simulation Setup} All distributed learning algorithms and server-clients connection are simulated using
the provided FedDCT training framework built on Pytorch
(version 1.10.2) available at: https://github.com/vinuni-vishc/fedDCT. We currently do not limit the number of resources a particular client possesses and simply measure the training cost. We run our simulation on a powerful server machine with the following hardware specifications: 
\begin{itemize}
    \item CPU : Dual Intel(R) Xeon(R) Gold 6242 CPU @ 2.80GHz, 16 Cores, 32 Threads each with 3.90 GHz Max Turbo Frequency.
    \item GPU : 8x NVIDIA RTX 3090 with 24 GB VRAM, 10496 CUDA Cores per card.
    \item 768 GB DDR4 ECC RAM running at 3200 MHz.
    \item Operating System: Ubuntu 22.04 LTS
\end{itemize}

\textbf{Implementation details for CIFAR-10}. For the CIFAR-10 dataset, we use WRN-16-8~\cite{zagoruyko2016wide} as the baseline model and train FedDCT (S = 4) for 300 communication rounds. In the training stage, all images are fed into the network
with a standard size of $32 \times 32$ pixels and a batch-size of $128$. The initial learning rate was initially set to $0.1$ and then gradually decreased using cosine learning rate decay policy~\cite{loshchilov2016sgdr}. In addition, common data augmentation techniques including random crop, random flipping, normalization, random erasing, mixup, and RandAugment~\cite{cubuk2020randaugment} are used during the training stage. We then train FedAvg~\cite{mcmahan2017communication}, FedProx~\cite{li2020federated}, FedGKT\cite{he2020group} and SplitFed~\cite{thapa2022splitfed} using the same training setting as FedDCT. All networks are trained with a $100\%$ participation rate of $20$ clients.

\textbf{Implementation details for CIFAR-100}. For the CIFAR-100 dataset, we use ResNet-110~\cite{he2016deep} as the baseline model and train FedDCT (S = 4) for $650$ communication rounds. The network is trained on $32 \times 32$ images with a batch size of $128$. The learning rate scheduler and data augmentation are similar to that used in the CIFAR-10 dataset. Also, FedAvg~\cite{mcmahan2017communication}, FedProx~\cite{li2020federated}, FedGKT\cite{he2020group} and SplitFed~\cite{thapa2022splitfed} then are trained on CIFAR-100 dataset using the same training procedures as the original works~\cite{mcmahan2017communication,li2020federated,thapa2022splitfed}.

\textbf{Implementation details for HAM10000}. For the HAM10000 dataset, we use WRN-50-2~\cite{zagoruyko2016wide} as the baseline model and train FedDCT (S = 4) for $200$ communication rounds. The network is trained on $64 \times 64$ images with a batch size of $128$. To reduce overfitting, we use different data augmentation strategies such as random vertical flip, random horizontal flip, random adjust sharpness, random auto contrast, rotation, and center crop. The initial learning rate is initially set to $0.01$ and then decreases by the cosine learning rate decay policy~\cite{loshchilov2016sgdr}. Similarly, we train FedAvg~\cite{mcmahan2017communication}, FedProx~\cite{li2020federated}, FedGKT\cite{he2020group} and SplitFed~\cite{thapa2022splitfed} for $200$ communication rounds using the same training setting as the original works~\cite{mcmahan2017communication,li2020federated,thapa2022splitfed, he2020group}.

\textbf{Implementation details for VAIPE}. For the VAIPE dataset, we also use WRN-50-2~\cite{zagoruyko2016wide} as the baseline model and then train the proposed FedDCT (S = 4) for $350$ communication rounds. The proposed framework is trained on pill images with the dimension of $256 \times 256$ pixels, crop size $224 \times 224$ pixels and with a batch size of $32$ samples. Data augmentation strategies on the VAIPE dataset are similar to that used in the CIFAR-10, and CIFAR-100 datasets, except that the label smoothing technique is applied. For FedAvg~\cite{mcmahan2017communication}, FedProx~\cite{li2020federated2}, SplitFed~\cite{thapa2022splitfed} and FedGKT\cite{he2020group}, we use $20$ clients with $100\%$ participation rate on IID data.
All other settings are the same as described in the original works~\cite{mcmahan2017communication,li2020federated,thapa2022splitfed,he2020group}.

\subsection{Accuracy evaluation} 
We report in this section the classification performance of FedDCT (S = 4) on CIFAR-10, CIFAR-100, HAM10000, and VAIPE datasets, respectively. We then compare our approach with recent state-of-the-art FL approaches, i.e., FedAvg~\cite{mcmahan2017communication}, FedProx~\cite{li2020federated2}, FedGKT~\cite{he2020group} and SplitFed~\cite{thapa2022splitfed}. We also provide a comprehensive analysis of the convergence of FedDCT to highlight the effectiveness of our method in terms of reducing the client’s memory requirements and communication costs.

\subsubsection{Model acccuracy}

Table~\ref{table:results} shows quantitative results of the proposed framework on CIFAR-10, CIFAR-100, HAM10000, and VAIPE datasets. Experimental results show that FedDCT is able to train large deep CNN models and reach a high level of performance. For instance, when using our FedDCT training strategy, the WRN-16-8 network achieves $95.99\%$ top-1 accuracy on the CIFAR-10 dataset. ResNet-100 achieves $77.38\%$ top-1 accuracy on CIFAR-100 dataset. Meanwhile, WRN-50-2 reports a top-1 accuracy of $78.62\%$ and $97.75\%$ on HAM10000 and VAIPE datasets, respectively. We observe that high-performing results is demonstrated consistently on a variety of model architectures and datasets. Fig~\ref{fig:1} shows the top-1 accuracy of all learning models on four datasets over communication rounds.

\begin{table}[t]
    \caption{Top-1 accuracy (\%) of FedProx, FedAvg, SplitFed, FedGKT and the proposed FedDCT (S = 4) on the test sets of CIFAR-10, CIFAR-100, HAM10000, and VAIPE datasets, respectively. Best results are marked in \textcolor{red}{\textbf{bold red}} and the second best are marked in \textcolor{blue}{\textbf{bold blue}}.}
    \label{table:results}
	\centering
	\scriptsize
	\resizebox{\linewidth}{!}{%
    \begin{tabular}{|l|l|c|l|l|l|l|}
        \hline \multirow{2}{*}{ \textbf{Dataset} } & \multirow{2}{*}{ \textbf{Model}} & \multicolumn{5}{|l|}{\hspace{1.5cm} \textbf{Top-1 Acc. (\%)} } \\
        \cline { 3 - 7 } & & \textbf{FedProx}& \textbf{FedAvg}& \textbf{SplitFed}& \textbf{FedGKT} & \textbf{FedDCT}\\
        \hline CIFAR-10 & WRN-16-8 & $89.02$ &  $92.12$ &  $91.98$ & \textcolor{blue}{\textbf{95.35}}& \textcolor{red}{\textbf{95.99}}   \\
        \hline CIFAR-100 & ResNet-110 & $66.92$ &  $73.67$ &  $72.92$ & \textcolor{blue}{\textbf{76.85}} & \textcolor{red}{\textbf{77.38}} \\
        \hline HAM10000 & WRN-50-2 & \textcolor{blue}{\textbf{78.05}} &  $77.13$ &  $76.14$&  $74.11$ &  \textcolor{red}{\textbf{78.62}} \\
        \hline VAIPE & WRN-50-2 & $86.20$ &  $88.79$ &  \textcolor{blue}{\textbf{91.85}} &  $56.68$ & \textcolor{red}{\textbf{97.75}} \\
        \hline
    \end{tabular}
    }
\end{table}




\subsubsection{Comparison with the state-of-the-arts}
Table~\ref{table:results} also provides the performance of the proposed model and previous methods on four image classification datasets. Concerning the CIFAR-10 dataset, FedDCT sets a new state-of-the-art performance with an accuracy of $95.99\%$. This result significantly outperforms other state-of-the-art FL method including FedGKT ($95.35\%$ top-1 accuracy), SplitFed ($91.98\%$ top-1 accuracy), FedAvg ($92.12\%$ top-1 accuracy), and FedProx ($89.02\%$ top-1 accuracy).  Compared to FedAvg, our FedDCT surpasses it by $3.87\%$ top-1 accuracy. On the CIFAR-100 dataset, FedDCT obtains a top-1 accuracy of $77.38\%$. Compared with FedProx, FedAvg, FedGKT and SplitFed, our method achieves much better performance, surpassing them by at least $+10.46\%$, $+3.71\%$, $+0.53\%$ and $+4.46\%$ in top-1 accuracy, respectively. On skin lesions dataset HAM10000, FedDCT reaches a top-1 accuracy of $78.62\%$ and significantly surpasses FedAvg, FedProx, FedGKT and SplitFed by $+1.49$,$+0.57\%$, $+4.51\%$ and $+2.48\%$, respectively. Concerning the pill image VAIPE dataset, FedDCT also outperforms all other state-of-the-art approaches and sets new state-of-the-art. Specifically, compared to the competing method SplitFed, our FedDCT significantly surpasses it by $+5.9\%$ top-1 accuracy while surpassing FedGKT, FedProx and FedAvg by $+41.07\%$, $+11.55\%$ and $+8.96\%$  top-1 accuracy, respectively. Experimental results on HAM10000 and VAIPE datasets suggest that the proposed FedDCT is a reliable FL approach to solving real-world applications.

\subsubsection{Convergence analysis}
\begin{table}[t]
    \caption{Number of communication rounds required to reach a target test-set accuracy and speedup relative to baseline FedAvg on the VAIPE dataset. Note that the symbol N/A (Not Available) means the method is not able to reach the target test accuracy.}
    \label{tab:2}
	\centering
	\scriptsize
	\setlength\tabcolsep{4pt} 
	\resizebox{0.85\linewidth}{!}{%
        \begin{tabular}{|l|r|r|l|l|l|}
            \hline \textbf{Top-1 Acc.} & \textbf{80 (\%)}  & \textbf{85 (\%)} & \textbf{90 (\%)} & \textbf{95 (\%)} \\
            \hline FedAvg & 195 (\texttt{base}) & 260 (\texttt{base}) & N/A  & N/A  \\
            \hline FedProx & 221 (\texttt{1.3$\times$}) & 285 (\texttt{1.1$\times$}) & N/A  & N/A  \\
            \hline SplitFed & 138 (\texttt{0.7$\times$}) & 177 (\texttt{0.7$\times$}) & 256  & N/A  \\
            \hline FedGKT & N/A & N/A & N/A  & N/A  \\
            \hline FedDCT (S=4) & 67 (\texttt{0.3$\times$}) & 86 (\texttt{0.3$\times$}) & 116  & 159  \\
            \hline
        \end{tabular}
    }
\end{table}
To demonstrate the effectiveness of the proposed FedDCT framework in reducing local updates, we provide the number of communication rounds and speedup relative of our method to the baseline FL method to reach a target test-set accuracy on the VAIPE dataset. We also compare the speed of convergence of FedDCT with other state-of-the-art FL methods. For the definition of a communication round of FedDCT, see section \ref{subsec:co-training} and \ref{subsec:clusteragg}. As shown in Table~\ref{tab:2}, FedAvg achieves a test accuracy of 80\% on the VAIPE dataset after 195 communication rounds (\texttt{base} speed). Meanwhile, the proposed approach FedDCT achieves a similar test accuracy after only 67 communication rounds (0.3$\times$\texttt{base}). This communication cost is much cheaper than FedProx and SplitFed, which need 221 (1.3$\times$\texttt{base}) rounds and 138 (0.7$\times$\texttt{base}) rounds, respectively. Besides, FedDCT reaches an accuracy of 85\% in only 86 communication rounds. Meanwhile, FedAvg needs 260 communication rounds. Table~\ref{tab:2} also shows that FedDCT converges much faster than existing approaches such as FedProx and SplitFed. We observed the same convergence behaviors to reach 85\%, 90\%, and 95\% on the test data of VAIPE. FedGKT is not able to converge to target accuracy on VAIPE dataset. In conclusion, our experimental results indicate that the proposed FedDCT requires fewer communication rounds to achieve the same level of accuracy compared to other state-of-the-art techniques.

This superiority in accuracy and convergence rate can be attributed to two things. First, recent research has demonstrated that an ensemble model can outperform a single model with the same number of parameters\cite{kondratyuk2020ensembling} and collaborative training between sub-models improves ensemble performance (see Section \ref{subsubsec:cotrainingcomp}). Similarly, FedDCT outperforms FedAvg with the same computation cost each communication round (despite training on more views of data, each view is trained on only a sub-model, hence a similar total training cost). Furthermore, due to the fact that each cluster of $S$ clients only trains a single ensemble model, the number of models in the aggregation phase is greatly decreased, hence lowering accuracy loss due to aggregation.

\subsection{Efficiency evaluation}
\begin{figure}[]
\centering
\vspace{-0.4cm}
\includegraphics[width=\linewidth]{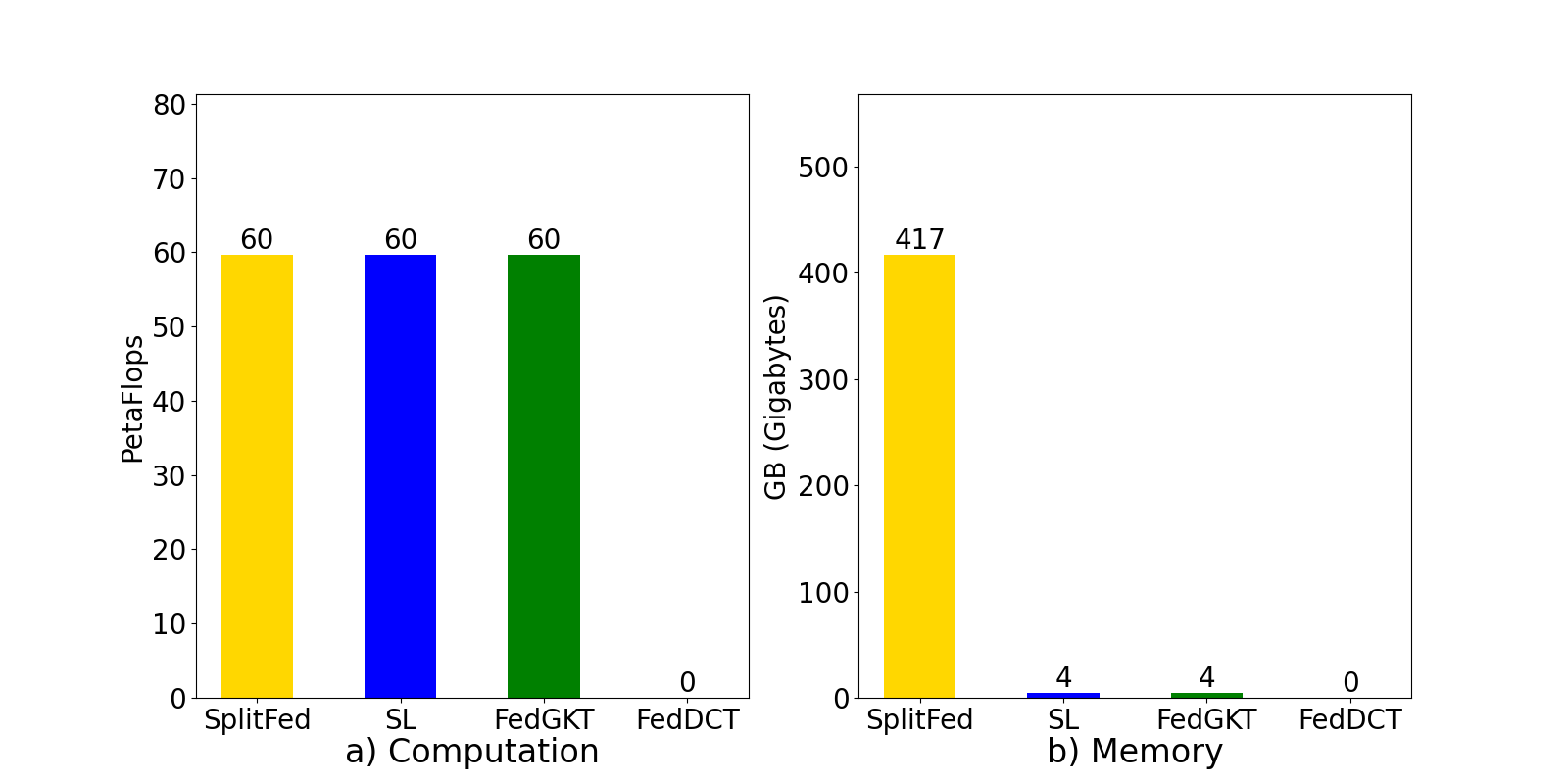}
\vspace{-0.6cm}
\caption{Training overhead at the server when training Split Learning (SL), SplitFed, FedGKT and FedDCT when training CIFAR-100 using ResNet-110 model: (a) the total amount of computation (petaFlops) for 650 communication rounds. (b) the required memory (GB) for training and aggregation with batch size 128, and 100 clients.}
\label{fig:server_cost}
\end{figure}

\subsubsection{Computation cost}
FedGKT\cite{he2020group}, Split Learning \cite{poirot2019split}, and SplitFed \cite{thapa2022splitfed} reduce the client's computation by offloading the majority of the training workload to the server side. All of these approaches assume that the training cost on the server side is negligible. While this might be true for laboratory settings where the dataset size is small, recent trends indicate that real-world datasets may be exceedingly huge\cite{narayanan2020analysis}, making training them extremely expensive. For example, a single training run of the GPT-3 \cite{brown2020language} model will take 355 V100 GPU-years and cost 4.6 million dollars.  To evaluate the computational cost caused by the training process at the server-side, we estimate the number of FLOPs (floating-point operations) required by our method and compare it to those of other approaches using the method given in \cite{OpenAI}. We report the results on the CIFAR 100 dataset with $60000$ images of the size $32 \times 32$ with $650$ communication rounds in Fig.~\ref{fig:server_cost}(a). As shown, compared to FedGKT, SplitFed and Split Learning, our method requires no training overhead on the server side.

Furthermore, in the case of SplitFed, where the training on the server side is run in parallel, the amount of VRAM required at the server increases linearly with the number of clients. In particular, SplitFed would require over 400GB of memory while training a ResNet-110 model on CIFAR-100 with batch size of $128$ using $100$ clients. Compared to other methods, FedDCT does not require the server to have GPUs (see Fig.~\ref{fig:server_cost}(b)).
\subsubsection{Client memory requirement}
\begin{figure}[]
\centering
\vspace{-0.2cm}
\includegraphics[width=1\linewidth]{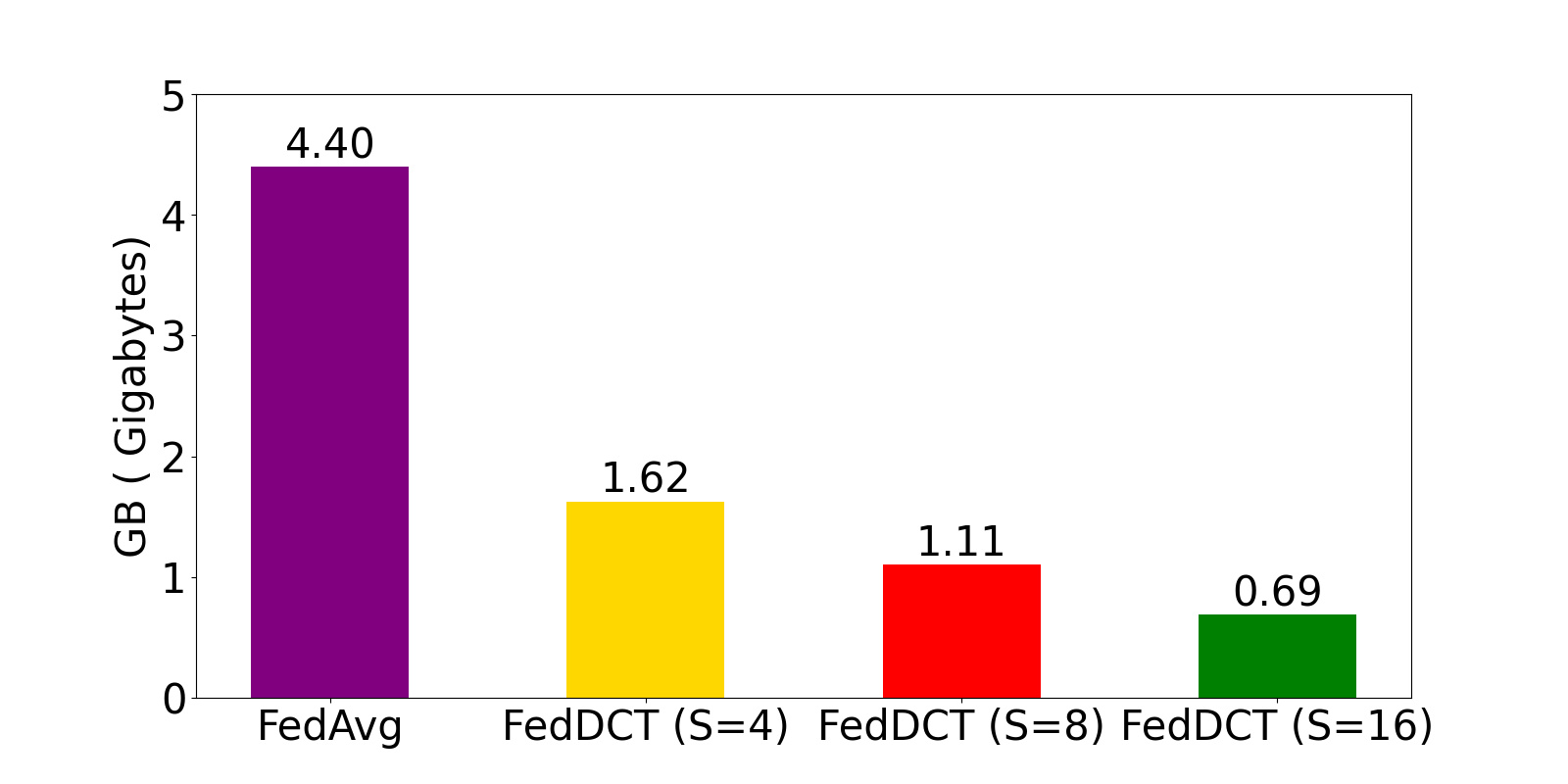} 
\vspace{-0.3cm}
\caption{Memory consumption of a client during training ResNet-110 on CIFAR-100 with FedDCT when the split factors $S$ are changed. FedAvg is equivalent to the case of $S=1$.} 
\label{fig:client_mem}
\end{figure}
\begin{figure}[h]
\centering
\includegraphics[width=\linewidth]{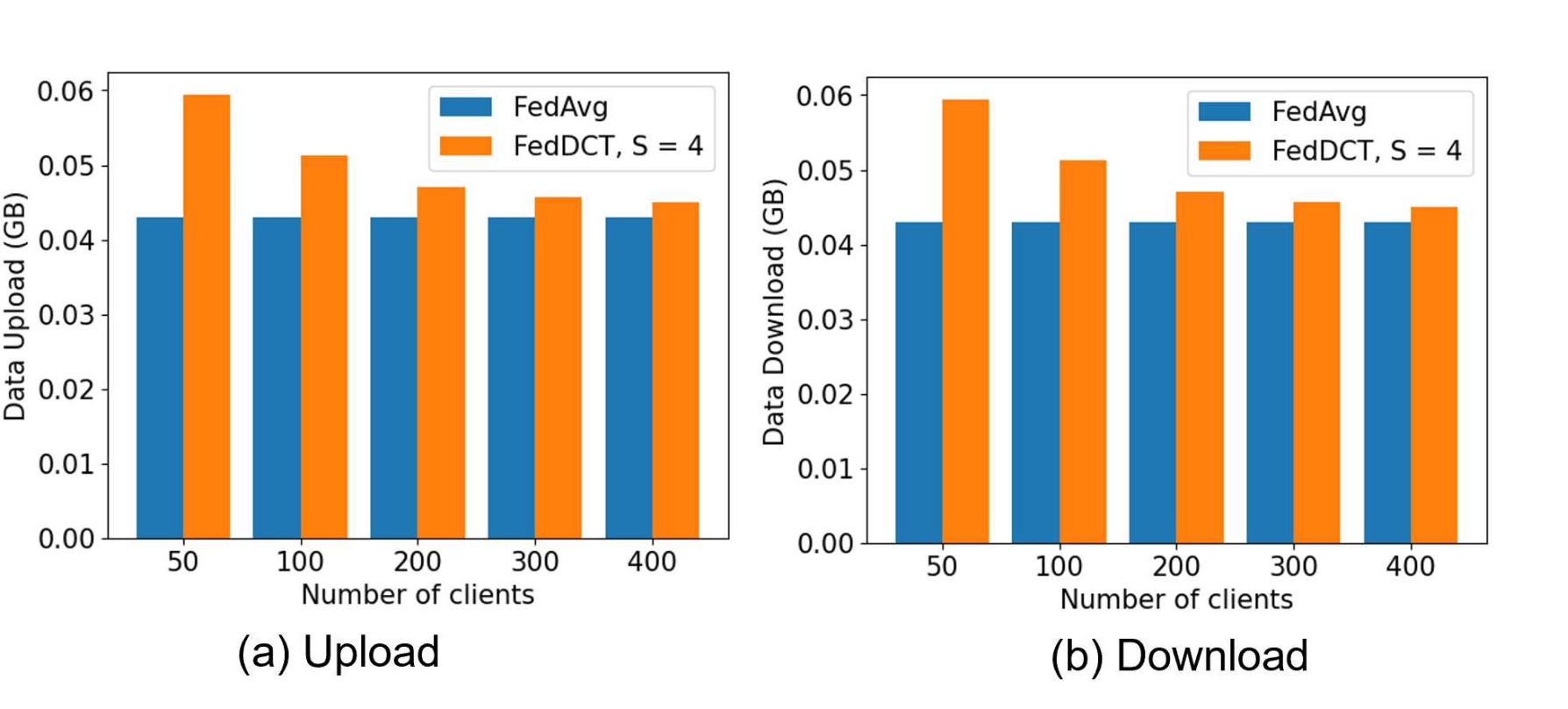}
\vspace{-0.6cm}
\caption{Communication measurement: average data per client per global epoch of FedAvg and FedDCT with split factor = 4 (a) upload for ResNet18 on HAM10000, (b)
download for ResNet18 on HAM10000.}
\label{fig:communication}
\end{figure}
\begin{figure*}
	\centering
	\begin{minipage}[!t]{0.32\textwidth}
        \includegraphics[width=\columnwidth, trim=0 0.5cm 0 1cm]{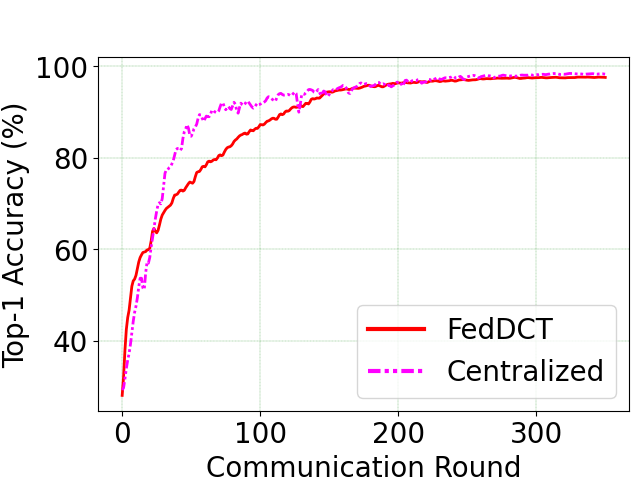}
        \caption{Convergence curves of centralized training approach and FedDCT on VAIPE dataset.}
        \label{fig:conver}
 	\end{minipage}
	\hspace{3pt}
	\begin{minipage}{0.32\textwidth}
        \includegraphics[width=\columnwidth, trim=0 0.5cm 0 1cm]{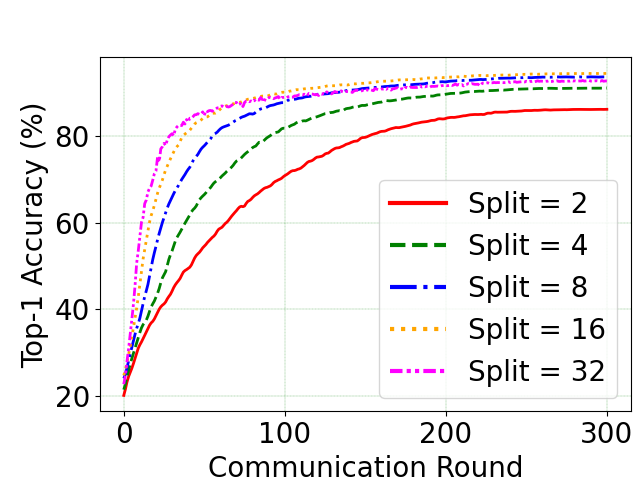}
        \caption{Effect of various number of splits  on the performance of the FedDCT on CIFAR-10 dataset.}
        \label{fig:splits}
 	\end{minipage}
	\hspace{3pt}
	\begin{minipage}{0.32\textwidth}
        \includegraphics[width=\columnwidth, trim=0 0.5cm 0 1cm]{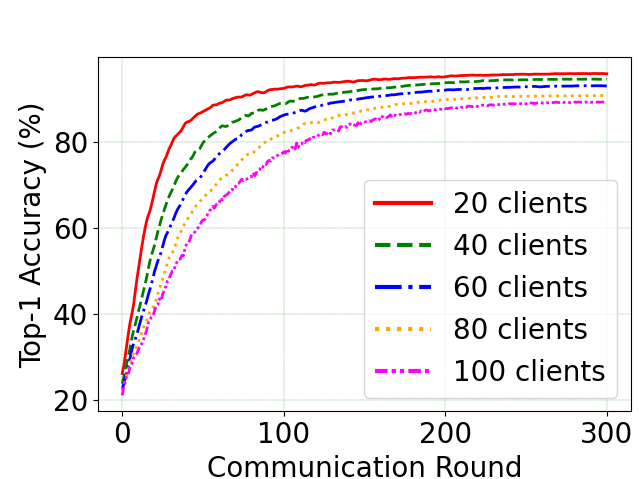}
        \caption{Effect of various number clients on the performance of FedDCT on CIFAR-10 dataset. }
        \label{fig:effect_clients}
 	\end{minipage}
\end{figure*}
This section studies the memory requirements of FedDCT under various settings of the split factor $S$, and compares them to that of FedAvg. 
As demonstrated in Fig.~\ref{fig:client_mem}, the greater the number of the split factor, the less memory is required to train the model on-device. Experiments on CIFAR-100 using Resnet-110 model with a batch size of $128$ demonstrates that while training the entire model on clients using FedAvg would consume $\approx 4.4$ GB memory, FedDCT consumes substantially less, for example, less than 1GB when $S=16$.

\subsubsection{Communication cost}
Let $K$ be the number of clients, $p$ be the total data size of all clients, $Q$ be the size of the smashed layer, $|{W}|$ be the size of the entire model, and $\beta$ be the ratio of the lower part to the entire model, i.e., $\left|{W}^{\mathrm{m}}\right|=\beta|{W}|$.
Note that, for each complete forward and backward propagation in FedDCT, a client may take one among two roles: main client and proxy client. 
The main client is in charge of generating the smashed data and performing a local model update on both lower and upper sub-models. In the meanwhile, the proxy clients receive the smashed data from the main client and perform the model update on only the upper sub-model. 
Accordingly, the communication cost for the owner of the training sample during each training iteration can be calculated as follows:
\begin{equation}
\small
\text{Comm. (main)}:(S-1)(\frac{2 p}{K})(\frac{Q}{S})+2 \beta|{W}|+2\frac{(1-\beta)|{W}|}{S}.
\end{equation}
On the other hand, the communication cost for each proxy client is determined as follows:
\begin{equation}
\small
\text{Comm. (proxy)}: \left(\frac{2 p}{K}\right) (\frac{Q}{S})+2(1-\beta) \frac{|{W}|}{S}.
\end{equation}
During each communication round, each client will be the main client one time and proxy client for $S-1$ times. Consequently, the total communication cost of each client in one communication round is  
\begin{equation}
\small
\text{Comm. (total)}: (S-1)(\frac{4p}{K})(\frac{Q}{S}) + 2|{W}|.
\end{equation}
The result of the communication efficiency of FedDCT compared to FedAvg is presented in Fig.~\ref{fig:communication}. As shown in the figure, FedDCT has a slight communication overhead per epoch compared to FedAvg due to the exchange of  smashed data and gradients between clients, but this overhead is not noticeable with a large number of clients. Furthermore, as we have observed in Table \ref{tab:2}, FedDCT achieves targeted accuracies much faster compared to FedAvg, hence requiring less total communication round, reducing total communication cost.

\subsection{Ablation studies}
In this section, we further conduct additional ablation studies to investigate the critical properties of the proposed FedDCT. Concretely, we conduct a performance comparison between FedDCT and centralized training. We then study the influence of the number of splits on learning performance. In addition, we also study the performance change over the number of clients, on non-IID datasets and the impact of collaborative training. We report ablation study results on CIFAR-10 and VAIPE datasets.

\subsubsection{Comparison to centralized training}

To further investigate the learning performance of FedDCT, we compare the proposed method and a centralized data training method, where the training data is gathered, and the entire training process executes at the central server~\cite{asad2021federated} on four datasets: CIFAR-10, CIFAR-100, HAM10000 and VAIPE. In particular, we train WRN-16-8, ResNet110 and WRN-50-2 on training examples of these datasets in a centralized manner. We report the test sets' performance and compare the centralized training approach and FedDCT (S=4) in Table~\ref{table:fed_vs_center}. Specifically, FedDCT (S=4) achieves 95.99\% top-1 accuracy on CIFAR-10 and 97.75\% on  VAIPE. Meanwhile, the centralized training models report an accuracy of 97.31\% on CIFAR-10 and 98.58\% on VAIPE. These results indicate that our FedDCT reaches almost the sample level of performance compared to centralized training in certain scenarios. For more difficult datasets such as CIFAR 100 or HAM10000, our method achieves a respectable 77.38\% and 78.62\%, respectively. Fig.~\ref{fig:conver} shows the convergence curves of FedDCT and the centralized training approach on the VAIPE dataset.

\begin{table}[t]
    \caption{Comparison of centralized training approach and FedDCT}    
    \begin{tabular}{|l|l|l|l|}
    \hline \textbf{Dataset} & \textbf{Model} & \multicolumn{2}{|c|}{\textbf{Top-1 Acc. (\%)}} \\
    \cline { 3 - 4 } & & \textbf{Centralized training} & \textbf{FedDCT (ours)} \\
    \hline CIFAR-10 & WRN-16-8 & \hspace{1cm}  $97.31$ & \hspace{0.6cm} $95.99$ \\
    \hline CIFAR100 & ResNet110 & \hspace{1cm}  $81.81$ & \hspace{0.6cm} $77.38$ \\
    \hline HAM10000 & WRN 50-2 & \hspace{1cm}  $83.92$ & \hspace{0.6cm} $78.62$ \\
    \hline VAIPE & WRN-50-2 & \hspace{1cm} $98.58$ & \hspace{0.6cm} $97.75$ \\
    \hline
    \end{tabular}
    \label{table:fed_vs_center}
\end{table}

\subsubsection{Effect of number of splits}

\begin{table}[t]
    \caption{Effect of the number of splits on Top-1 Accuracy of FedDCT}  
    \begin{tabular}{|l|l|ccccc|}
    \hline
    \multirow{2}{*}{\textbf{Dataset}} & \multirow{2}{*}{\textbf{Model}} & \multicolumn{5}{c|}{\textbf{Number of splits}}                                                                                                       \\ \cline{3-7} 
                                      &                                 & \multicolumn{1}{c|}{\textbf{2}} & \multicolumn{1}{c|}{\textbf{4}} & \multicolumn{1}{c|}{\textbf{8}} & \multicolumn{1}{c|}{\textbf{16}} & \textbf{32} \\ \hline
    \multicolumn{1}{|c|}{CIFAR 10}    & \multicolumn{1}{c|}{WRN 16-8}   & \multicolumn{1}{c|}{86.21}          & \multicolumn{1}{c|}{91.13}          & \multicolumn{1}{c|}{93.74}          & \multicolumn{1}{c|}{94.52}           & 92.87           \\ \hline
    \multicolumn{1}{|c|}{CIFAR 100}   & \multicolumn{1}{c|}{WRN 16-8} & \multicolumn{1}{c|}{73.48}          & \multicolumn{1}{c|}{77.91}          & \multicolumn{1}{c|}{79.54}          & \multicolumn{1}{c|}{78.57}           & 73.19         \\ \hline
    \end{tabular}
    \label{table:numsplits}
\end{table}
In this experiment, we investigate the effect of the number of splits on the performance of FedDCT. The experiment is conducted on the CIFAR-10 and CIFAR-100 datasets, with the number of splits is set at 2, 4, 8, 16, and 32, respectively. To this end, we use WRN-16-8 as the baseline network. The experimental settings and training methodology are similar to our main experiments described in \textcolor{blue}{Section~\ref{implementation_details}}. Learning behaviors of FedDCT with a different number of splits are shown in Fig.~\ref{fig:splits} and Table \ref{table:numsplits}. We find that the WRN-16-8 achieves the highest top-1 accuracy with the number of splits equal to 16 for CIFAR-10 and 8 for CIFAR-100, with the performance of FedDCT continuously increases when the number of splits increases from 2 to 8. However, a significant drop in performance is observed when we increase the number of splits to 32. Explaination: The sub-networks may be too thin to guarantee sufficient model capacity.

\subsubsection{Effect of number clients}
\begin{table}[t]
    \caption{Effect of the number of clients on Top-1 Accuracy of FedDCT}  
    \begin{tabular}{|l|l|ccccc|}
    \hline
    \multirow{2}{*}{\textbf{Dataset}} & \multirow{2}{*}{\textbf{Model}} & \multicolumn{5}{c|}{\textbf{Number of clients}}                                                                                                       \\ \cline{3-7} 
                                      &                                 & \multicolumn{1}{c|}{\textbf{20}} & \multicolumn{1}{c|}{\textbf{40}} & \multicolumn{1}{c|}{\textbf{60}} & \multicolumn{1}{c|}{\textbf{80}} & \textbf{100} \\ \hline
    \multicolumn{1}{|c|}{CIFAR 10}    & \multicolumn{1}{c|}{WRN 16-8}   & \multicolumn{1}{c|}{95.99}          & \multicolumn{1}{c|}{94.55}          & \multicolumn{1}{c|}{93.02}          & \multicolumn{1}{c|}{90.75}           & 89.21           \\ \hline
    \multicolumn{1}{|c|}{CIFAR 100}   & \multicolumn{1}{c|}{ResNet110} & \multicolumn{1}{c|}{77.38}          & \multicolumn{1}{c|}{73.94}          & \multicolumn{1}{c|}{71.55}          & \multicolumn{1}{c|}{70.02}           & 68.84         \\ \hline
    \end{tabular}
    \label{table:numsclients}
\end{table}
We study the effect of the number of clients on the performance of FedDCT. The proposed method is trained and evaluated on the CIFAR-10 dataset with a number of clients of 20, 40, 60, 80, and 100, respectively. Fig.~\ref{fig:effect_clients} and Table \ref{table:numsclients} shows the effect of various number clients on FedDCT (S=4) performance on CIFAR-10 and CIFAR-100 datasets with WRN-16-8 and ResNet 110 as the baseline model, respectively. We found that the use of 20 clients allows FedDCT to achieve the highest top-1 accuracy (95.99\%) for CIFAR 10 and 77.38\% for CIFAR-100. We also observe that the model's performance decreases when the number of clients increases. Our main observation is that usually, the convergence slows down, and performance degrades with the increase in the number of users within the observation window of the global epochs. Furthermore, all our DCML approaches show similar behavior over the various number of users (clients). This behavior is also observed in other FL techniques \cite{mcmahan2017communication,thapa2022splitfed,poirot2019split}.
\subsubsection{Non-IID datasets} 
\begin{figure*}[t]
\centering
\vspace{-0.5cm}
\includegraphics[width=1\linewidth]{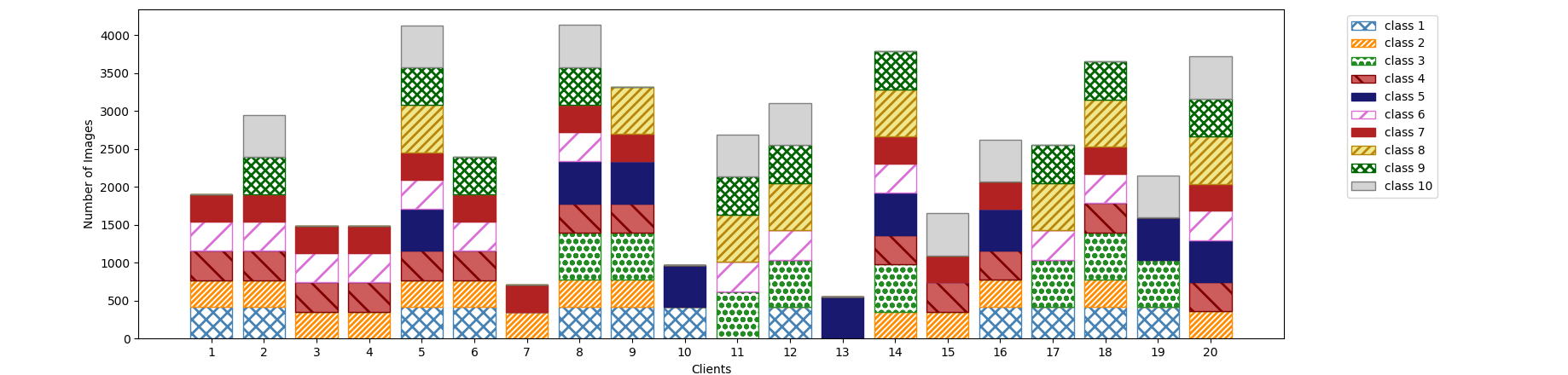} 
\caption{Non-IID data distribution of 20 clients for CIFAR-10 dataset. CIFAR-100 is divided using a similar partition scheme.} 
\label{fig:noniid}
\end{figure*}

\begin{figure}[t]
\centering
\vspace{-0.5cm}
\includegraphics[width=0.8\linewidth]{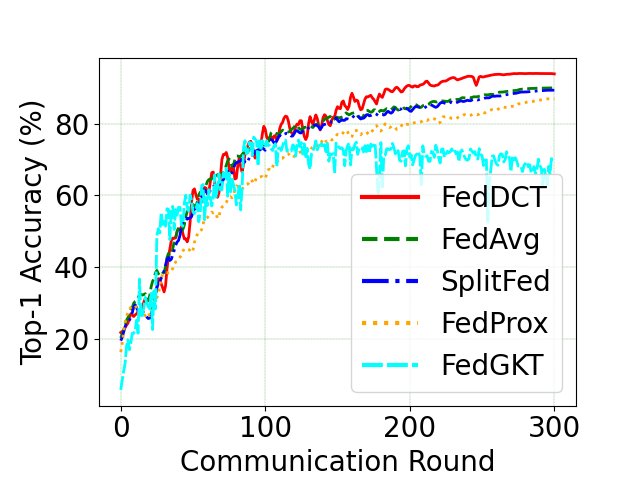} 
\caption{Top-1 accuracy (\%) of FedDCT (S=4) compared to state-of-the-art FL methods on the test sets of non-IID CIFAR-10 dataset.} 
\label{fig:noniidgraph}
\end{figure}

To create a non-IID dataset, CIFAR 10 and CIFAR 100 are randomly divided into the given number of clients (see Fig. \ref{fig:noniid}), where each client can have a different number of classes and samples. This type of dataset replicates the real-world scenario where clients can have different types of images.
We report the test sets' performance of FedDCT (S=4) compared to FedAvg, SplitFed, FedGKT and FedProx in Fig.~\ref{fig:noniidgraph}. In non-IID settings, FedDCT shows superior performance compared to other methods, i.e., achieving 94.01\% top-1 accuracy with WRN-16-8 as the baseline model, and 75.16\% top-1 accuracy on CIFAR-100 with ResNet-110 as the baseline model. We observe that while FedDCT might experience slight fluctuation initially during training compared to other methods, it tends to stabilize during the latter half of the training process when the model is trained with enough diversity in data. Our theory is that due to the sequential selection of the main client, the ensemble model will only train with the main client's data until a new main client is selected and hence the updated model will be biased to the main client's data. To solve this, a more advanced data selection technique can be used, where during each forward pass, a client will be selected at random to be the new main client.
\begin{figure}[t]
\centering
\vspace{-0.5cm}
\includegraphics[width=0.8\linewidth]{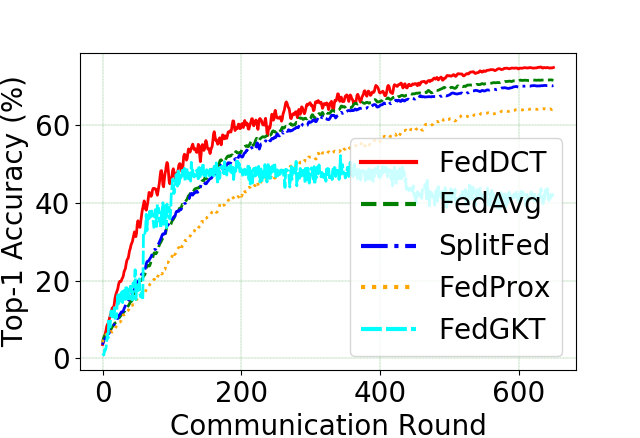} 
\caption{Top-1 accuracy (\%) of FedDCT (S=4) compared to state-of-the-art FL methods on the test sets of non-IID CIFAR-100 dataset.} 
\label{fig:noniidcf100}
\end{figure}
\subsubsection{Influences of different collaborative training components}
\label{subsubsec:cotrainingcomp}
\begin{table}[t]
    \caption{Impact of different collaborative training components on the performance of FedDCT (S =4) on CIFAR-10 using WRN 16-8.}    
    \centering
    \begin{tabular}{|c|c|cccc|} 
    \hline
    Dataset  & Model    & diff. init & diff. views & $\lambda_{cot}$ & Top-1 Acc.(\%)  \\ 
    \hline
             &          & \xmark          & \xmark           &       &  94.88          \\
             &          & \cmark          & \xmark           &       & 95.01 \\
             &          & \cmark          & \cmark           &       & 95.31           \\
    CIFAR-10 & WRN  & \cmark          & \cmark           & 0.1       & 95.68                \\
             & 16-8                   & \cmark  & \cmark & 0.5   & 95.99        \\
             &          & \cmark          & \cmark           & 1.0   & 95.23           \\
    \hline
    \end{tabular}
    \label{table:co-training}
\end{table}
Table \ref{table:co-training} shows the impacts of varying weight initialization, various weight factor values of collaborative training loss in Equation \ref{eq:objectivefuncgeneral}, and various diverse data perspectives. Using different data transformers (0.30\%↑) and collaborative training loss (0.68\%↑) can help the model improve performance. Given the strong baseline, this improvement is also noteworthy. In this work, we simply apply a straightforward collaborative training strategy and don't further explore this topic. Other, more complicated collaborative training or mutual learning techniques do exist. For instance, to improve the performance of the entire network, MutualNet\cite{yang2020mutualnet} creates numerous networks with different widths from a single full network, feeds them images at varied resolutions, and trains them jointly in a weight-sharing manner. More elaborate collaborative training methods are left as future work since we primarily concentrate on demonstrating that collaborative training can help enhance overall performance.
\section{Discussion}
\label{discussion}
\subsection{Privacy Protection}
FedDCT maintains privacy through two key features. Firstly, it employs a model-to-data approach ( model, or in this case intermediate layer itself is transferred rather than data). Secondly, FedDCT employs a split network during ML training, allowing for the full privacy of main client and proxy client models. This is accomplished by preventing the one client from receiving model updates of another client and vice versa. Instead, the each proxy client only has access to the smashed data (activation vectors of the cut layer) of a sub-model of the main client. To infer main client model parameters and raw data, the proxy client must be able to eaves-drop the smashed data sent to all other proxy clients and invert all client-side model parameters, a highly unlikely possibility if the main client and other proxy clients' network is configured with sufficiently large numbers of \cite{gupta2018distributed}. However, smaller main client networks may be susceptible to this issue, which can be controlled by modifying the loss function at the client-side\cite{vepakomma2019reducing}. Main client are also unable to infer proxy clients model parameters, as they only have access to the gradients of the smashed data and main-client-side updates, respectively. Compared to FL, FedDCT provides superior architectural configurations for enhanced privacy during ML model training, as it employs a network split and separate training on the main-client-side and proxy-clients-side.\newline
Despite FedDCT's inherent privacy protection properties, there is still a risk of advanced adversaries exploiting information representations of shared data or parameter weights to violate data owner privacy. While our work does not address this privacy concern in detail, we believe that existing methods such as differential privacy (DP)\cite{abadi2016deep,lecuyer2019certified} and multi-party computation (MPC)\cite{kanagavelu2020two} can provide protection against hidden vector reconstruction attacks. It is important to note that gradient exchange occurs during the training phase, making the attack more challenging since the attacker's access is limited to the evolving and untrained feature map, rather than the fully trained feature map that represents the raw data. However, it is essential to consider that model and gradient exchange may also result in privacy leaks, and the degree of privacy leakages between these three settings (gradient, model, and hidden map) was not analyzed or compared in our work, representing a limitation. However, this is a problem faced by other comptemporary works \cite{poirot2019split,thapa2022splitfed,he2020group}, and this is an ongoing research field.
\subsection{Real-World Applications}
FedDCT presents a promising approach for efficient federated training of large deep neural networks on resource-constrained edge devices, such as smartphones, IoT devices, and edge servers. By leveraging numerous small client models, FedDCT demonstrates the feasibility of training substantial server-side models while upholding the privacy standards of Federated Learning (FL) within the confines of edge computing. This technique holds particular benefits for smartphone users, ensuring data privacy while offering high-quality model services. Furthermore, FedDCT facilitates collaborative training of large CNN models, proving advantageous for organizations with limited training resources like hospitals and non-profit entities. This collaborative approach distributed across multiple clients yields substantial cost savings. Additionally, FedDCT addresses concerns related to intellectual property protection, confidentiality, regulatory compliance, and legal constraints. In practical applications, FedDCT finds relevance in various domains such as healthcare, surveillance systems, and smart homes. For example, in healthcare, the approach allows model training using patient data spread across clients' devices, alleviating privacy concerns associated with centralized data storage. In surveillance systems, security providers can develop models for tasks like face recognition and people detection without compromising user data confidentiality. In the context of smart homes, FedDCT extends its utility to training models on low-power IoT devices deployed across households. Although FedDCT offers substantial benefits, it's important to acknowledge potential challenges and limitations associated with its implementation.

\section{Conclusion and Future Works}
\label{conclusion}
Recent works on FL motivated us to find a robust solution for training high-performing, large-scale deep neural networks in edge computing devices. This paper proposed FedDCT, a new approach for training large deep neural networks in FL settings. The proposed learning framework allows dividing large CNN networks into sub-networks and collaboratively train them with local clients in parallel. To the best of our knowledge, this is the first work that enables the training of large deep learning networks in edge computing devices in FL settings. Extensive experiments on various benchmarks show that the proposed FedDCT reduces the number of local updates compared with the state-of-the-art FL methods and makes a new state-of-the-art for the image classification task. We also showed that FedDCT could provide a higher level of performance over baseline protocols for training diverse machine learning models on real-world medical imaging data.

Our work faces a few limitations. First, the proposed method applies to CNN models. However, our dividing and collaborative training strategy are hard to apply for time series learning architectures such as recurrent neural networks~\cite{sherstinsky2020fundamentals} (RNN) or famous segmentation networks such as U-Net and its variants~\cite{siddique2021u}. Second, the slowest client determines the pace of forward propagation and backpropagation in a cluster. Nonetheless, this is an issue faced by FedAvg~\cite{mcmahan2017communication}, SplitFed~\cite{thapa2022splitfed}, and other variants~\cite{poirot2019split}. Several client selection methods such as \cite{nishio2019client,huang2022stochastic} have been proposed to solve this problem and can be incorporated into FedDCT, however, this is outside the scope of this work. By allowing a client to train only a portion of the model, we assume that the client will be able to complete the training process. Additionally, considering the heterogeneous resources between the devices, an interesting future work is to divide the global model into unequal partitions and assign the larger sub-model to more capable clients. Another interesting direction is to combine the proposed FedDCT framework with state-of-the-art robust aggregation protocols (e.g.,~\cite{wang2020federated,wang2020tackling}) or clustering strategies \cite{khan2021socially} while ensuring privacy. For instance, if intra-cluster aggregation is carried out first, and then the complete model is transmitted from the cluster head to the server, the communication efficiency may be improved, especially when some users in the cluster are far away from the server~\cite{lin2021semi}.

\section*{Acknowledgments}
This work was funded by Vingroup Joint Stock Company and supported by Vingroup Innovation Foundation (VINIF) under project code VINIF.2021.DA00128 and also supported by JSPS KAKENHI Grant Numbers JP21K17751. We thank all our collaborators who participated in the collection and annotation of the VAIPE dataset.

{\appendix[Details of Dividing Large Model]
\label{appendix}
Our dividing method is robust and can apply to most modern CNNs. In this section, we consider some of the most common architectures. Here $S$ is the split factor (the number of small networks after dividing).\\
\\
\textbf{ResNet} For CIFAR-10 and CIFAR-100 datasets, the input channel numbers of the three blocks are as follows
\begin{equation}
\begin{aligned}
\text { baseline : } & {[16,32,64], } \\
S=2: & {[12,24,48], } \\
S=4: & {[8,16,32] , } \\
S=8: & {[6,12,23], } \\
S=16: & {[4,8,16] , }\\
S=32: & {[3,6,12]. } 
\end{aligned}
\end{equation}
\textbf{Wide-ResNet} The input channel is divided similarly to ResNet. If the widen factor is $f_w$, the new widen factor after division is ${f_w}^{\star}$ 
\begin{equation}
{f_w}^{\star}=\max \left(\left\lfloor\frac{f_w}{\sqrt{S}}+0.4\right\rfloor, 1.0\right).
\end{equation}
\textbf{ResNeXt\cite{xie2017aggregated}} Assume the original cardinality (groups in convolution) is $f_{c}$ and the new cardinality is ${f_{c}}^{\star}$, in which
\begin{equation}
{f_{c}}^{\star}=\max \left(\left\lfloor\frac{f_{c}}{S}\right\rfloor, 1.0\right).
\end{equation}
\textbf{EfficientNet\cite{tan2019efficientnet}} The number of output channels and blocks in the first convolutional layer of the EfficientNet baseline are as follows
\begin{equation}
\begin{aligned}
\text { baseline : } & {[32,16,24,40,80,112,192,320,1280]}, \\
S=2: & {[24,12,16,24,56,80,136,224,920]}, \\
S=4: & {[16,12,16,20,40,56,96,160,640]}.
\end{aligned}
\end{equation}
\textbf{DenseNet\cite{huang2017densely}} If DenseNet's growth rate is $f_g$, then the new growth rate after division ${f_{g}}^{\star}$ is
\begin{equation}
{f_{g}}^{\star}=\frac{1}{2} \times\left\lfloor 2 \times \frac{f_g}{\sqrt{S}}\right\rfloor .
\end{equation}}


\bibliographystyle{IEEEtran}
\bibliography{references}

\section{Biography Section}
 

\begin{IEEEbiography}[{\includegraphics[width=1in,height=1.25in,clip,keepaspectratio]{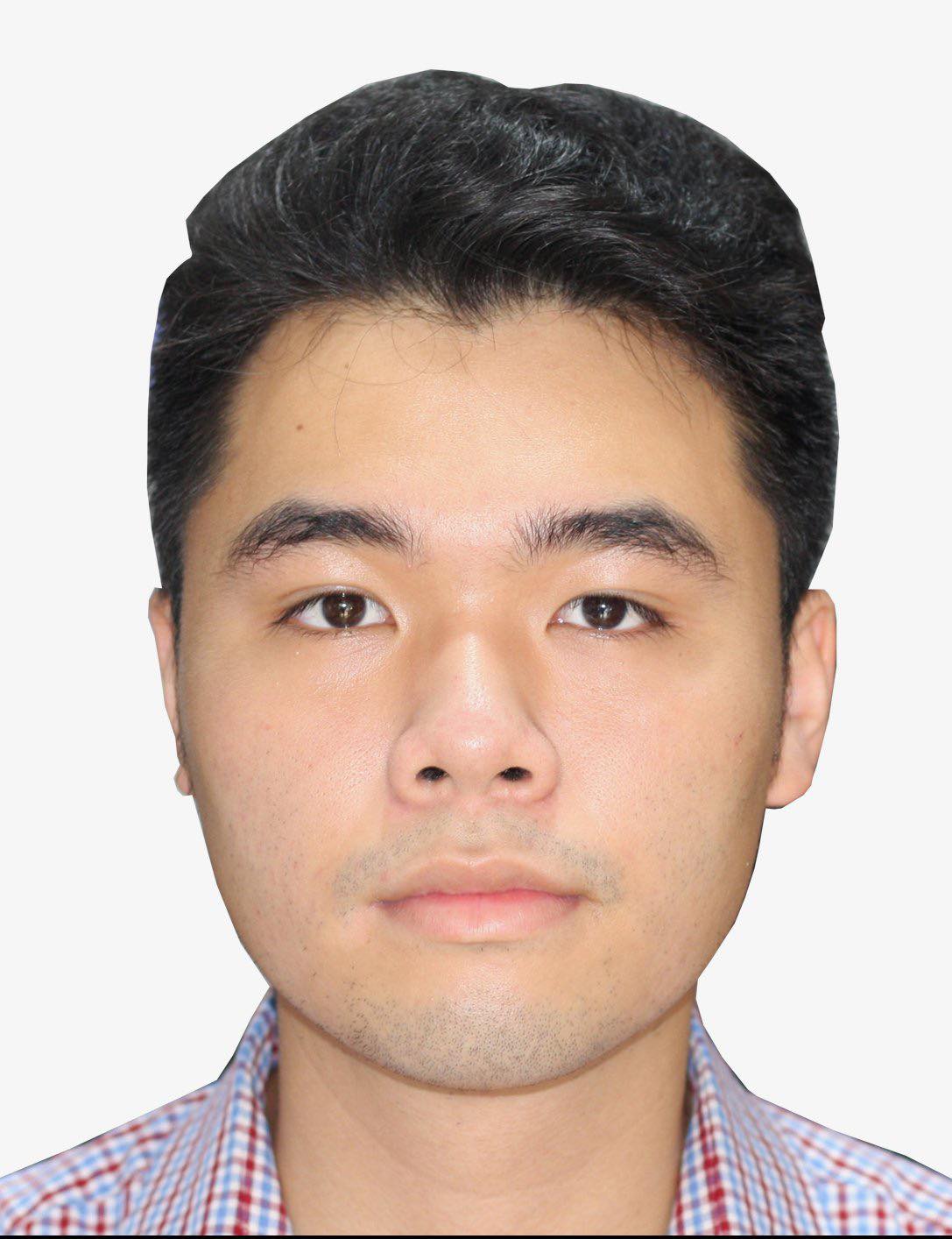}}]{Quan Nguyen} received his Engineering degree in Biomedical Engineering at Hanoi University of Science of Technology (HUST), Vietnam in 2021, and B.Sc. in Computer Science at HUST in 2022. He is currently a Master Student at Department of Informatics, Technical University Munich and Research Assistant at VinUni-Illinois Smart Health Center (VISHC). His research interests lies in the intersection between artificial intelligence and healthcare, including computer-aided medical procedures, inverse problems in medical imaging, federated learning and deep learning on the edge. Before joining VISHC, he worked as an AI engineer at Viettel High Technology Industries Corporation where he developed Computer Vision Algorithms for smart camera systems. He is the author of various scientific articles published in reputable conferences on surveillance systems such as IEEE International Conference on Advanced Video and Signal-Based Surveillance (AVSS). 
\end{IEEEbiography}
\begin{IEEEbiography}[{\includegraphics[width=1in,height=1.25in,clip,keepaspectratio]{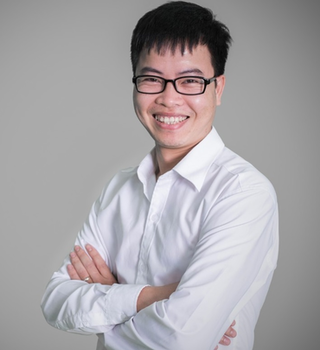}}]{Hieu Pham} (Member, IEEE) received the Engineering degree in Industrial Informatics from the Hanoi University of Science and Technology (HUST), Vietnam, in 2016, and the Ph.D. degree in Computer Science from the Toulouse Computer Science Research Institute (IRIT), University of Toulouse, France, in 2019. He is currently an Assistant Professor at the College of Engineering and Computer Science (CECS), VinUniversity, and serves as an Associate Director at the VinUni-Illinois Smart Health Center (VISHC). His research interests include computer vision, machine learning, medical image analysis, and their applications in smart healthcare. He is the author and co-author of 30 scientific articles appeared in about 20 conferences and journals, such as Computer Vision and Image Understanding, Neurocomputing, Scientific Data (Nature), International Conference on Medical Image Computing and Computer-Assisted Intervention (MICCAI), Medical Imaging with Deep Learning (MIDL), IEEE International Conference on Image Processing (ICIP), and IEEE International Conference on Computer Vision (ICCV), Asian Conference on Computer Vision (ACCV). He is also currently serving as Reviewer for MICCAI, ICCV, CVPR, IET Computer Vision Journal (IET-CVI), IEEE Journal of Biomedical and Health Informatics (JBHI), IEEE Journal of Selected Topics in Signal Processing, and Scientific Reports (Nature). Before joining VinUniversity, he worked at the Vingroup Big Data Institute (VinBigData), as a Research Scientist and the Head of the Fundamental Research Team. With this position, he led several research projects on medical AI, including collecting various types of medical data, managing and annotating data, and developing new AI solutions for medical analysis.
\end{IEEEbiography}
\begin{IEEEbiography}[{\includegraphics[width=1in,height=1.25in,clip,keepaspectratio]{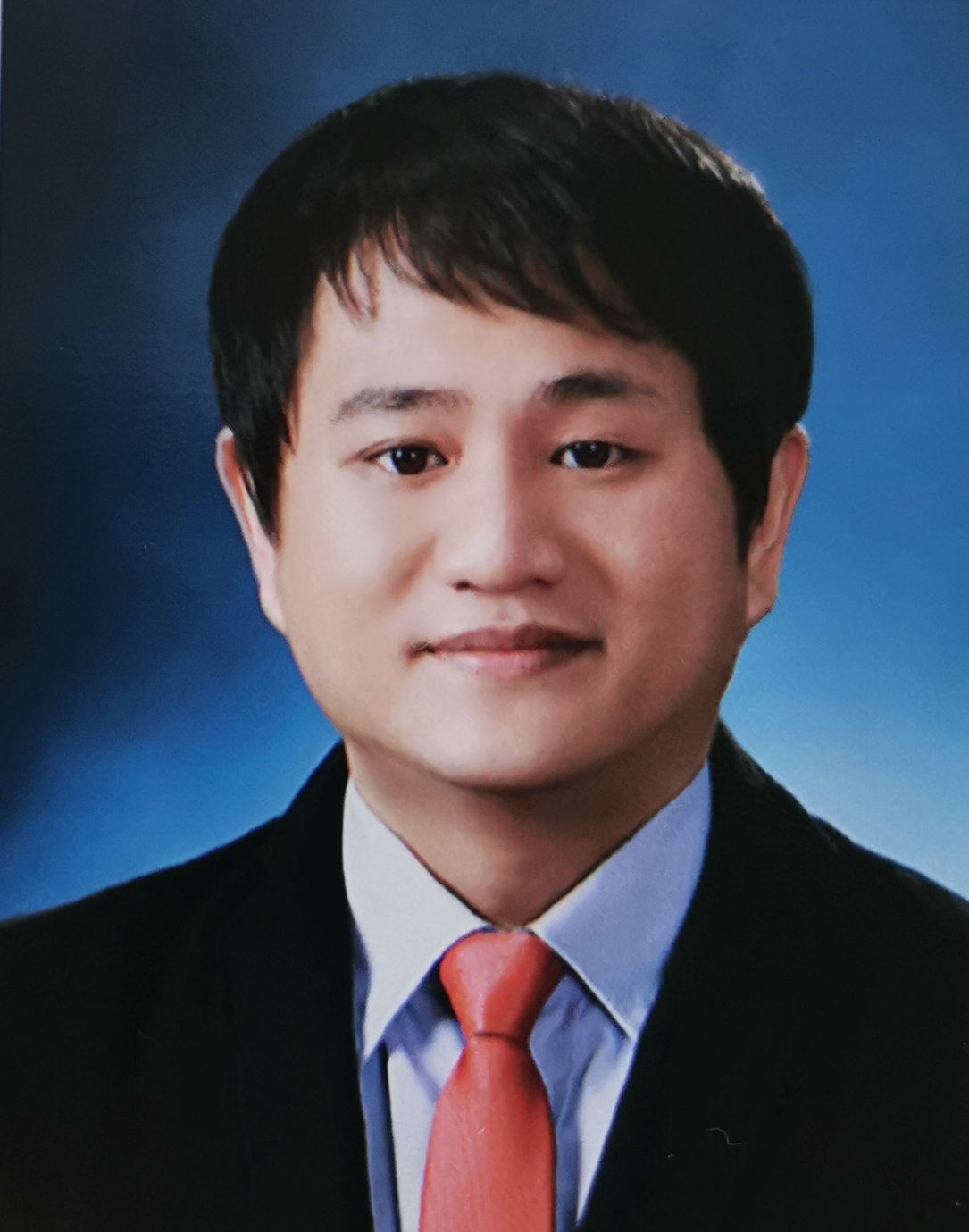}}]{Kok-Seng Wong}(Member, IEEE)
received his first degree in Computer Science (Software Engineering) from the University of Malaya, Malaysia in 2002, and an M.Sc. (Information Technology) degree from the Malaysia University of Science and Technology (in collaboration with MIT) in 2004. He obtained his Ph.D. from Soongsil University, South Korea, in 2012. He is currently an Associate Professor in the College of Engineering and Computer Science, VinUniversity. Before joining VinUniversity, he taught computer science subjects (undergraduate and postgraduate) in Kazakhstan, Malaysian, and South Korean universities for the past 17 years. Dr. Wong aims to bring principles and techniques from cryptography to the design and implementation of secure and privacy-protected systems. He has published 60 articles in journals and conferences in his research field. To this end, he conducts research that spans the areas of security, data privacy, and AI security while maintaining a strong relevance to the privacy-preserving framework. 
\end{IEEEbiography}
\begin{IEEEbiography}[{\includegraphics[width=1in,height=1.25in,clip,keepaspectratio]{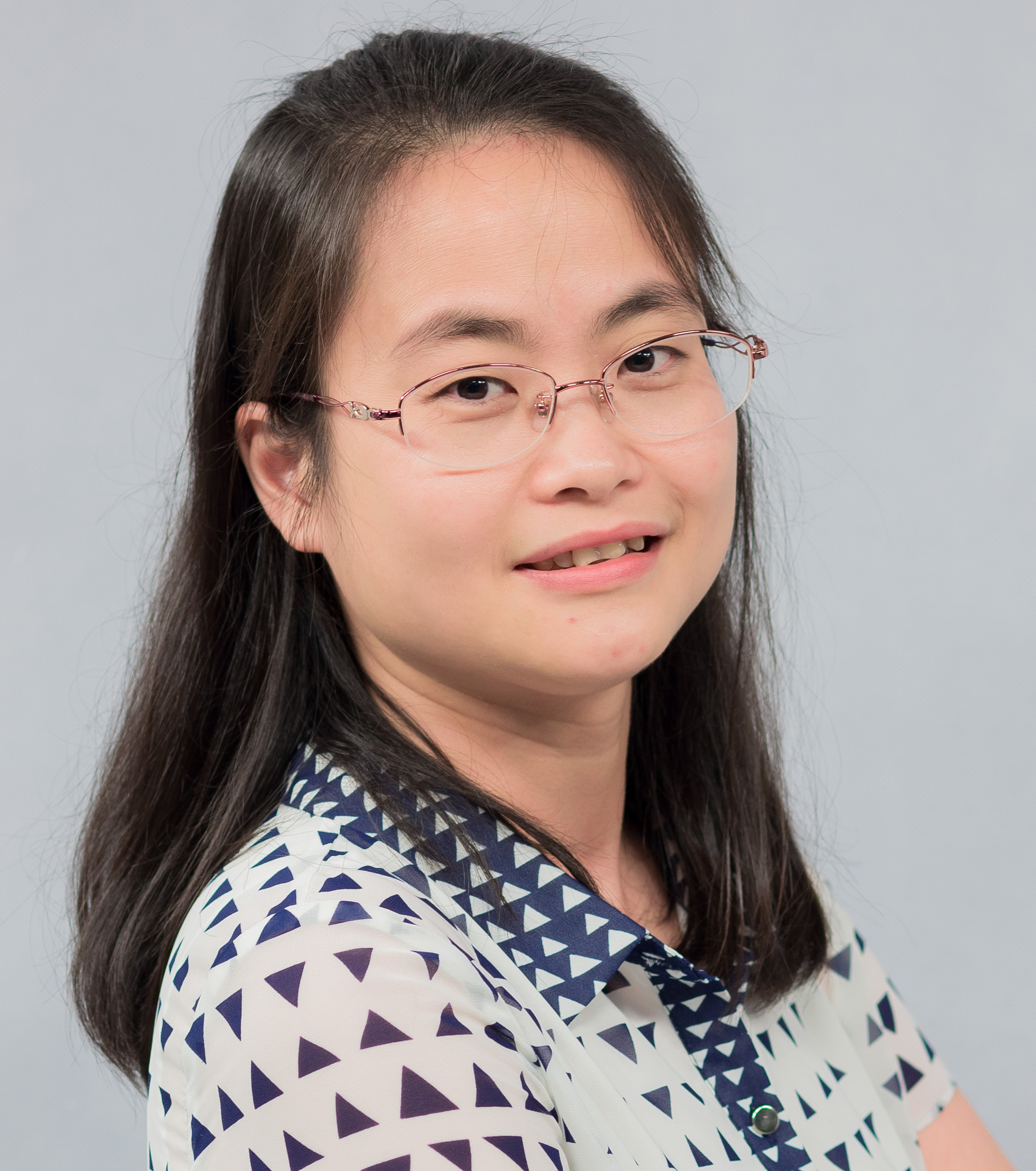}}]{Phi Le Nguyen}
received her B.E. and M.S. degrees from the University of Tokyo in 2007 and 2010, respectively. She received her Ph.D. in Informatics from The Graduate University for Advanced Studies, SOKENDAI, Tokyo, Japan, in 2019. Currently, she is a lecturer at the School of Information and Communication, Hanoi University of Science and Technology (HUST), Vietnam. In addition, she is serving as the managing director at the International research center for Artificial Intelligence (BKAI), HUST. Her research interests include Network architecture, Optimization, and Artificial Intelligence.
\end{IEEEbiography}
\begin{IEEEbiography}[{\includegraphics[width=1in,height=1.25in,clip,keepaspectratio]{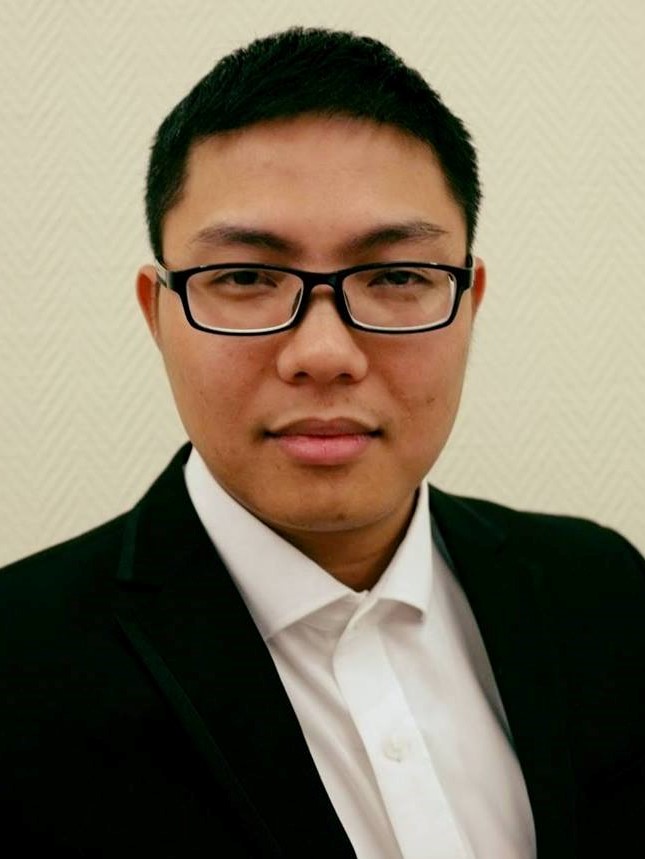}}]{Truong Thao Nguyen}
Dr. Truong Thao Nguyen received the BE and ME degrees from Hanoi University of Science and Technology, Hanoi, Vietnam, in 2011 and 2014, respectively. He received the Ph.D. in Informatics from the Graduate University for Advanced Studies, Japan in 2018. 
 He is currently working at 
 Digital Architecture Research Center, at National Institute of Advanced Industrial Science and Technology (AIST), where he focuses on the topics of High Performance Computing system, Distributed Deep Learning and beyond.
\end{IEEEbiography}
\begin{IEEEbiography}[{\includegraphics[width=1in,height=1.25in,clip,keepaspectratio]{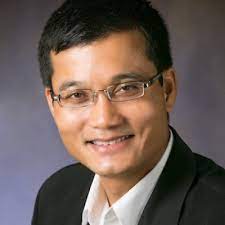}}]{Minh Do}
(Fellow, IEEE) was born in Vietnam in 1974. He received the B.Eng. degree in computer engineering from the University of Canberra, Australia, in 1997, and the Dr.Sci. degree in communication systems from the Swiss Federal Institute of Technology Lausanne (EPFL), Switzerland, in 2001. Since 2002, he has been on the Faculty of the University of Illinois at Urbana-Champaign (UIUC), where he is currently the Thomas and Margaret Huang Endowed Professor in signal processing and data science with the Department of Electrical and Computer Engineering, and holds affiliate appointments with the Coordinated Science Laboratory, Department of Bioengineering, and the Department of Computer Science, Beckman Institute for Advanced Science and Technology. From 2020 to 2021, he is on leave from UIUC to serve as the Vice Provost for VinUniversity, Vietnam.
Dr. Do was a member of several IEEE technical committees on signal processing. He was elected as a fellow of IEEE in 2014 for his contributions to image representation and computational imaging. He received the Silver Medal from the 32nd International Mathematical Olympiad in 1991, the University Medal from the University of Canberra in 1997, the Doctorate Award from the EPFL in 2001, the CAREER Award from the National Science Foundation in 2003, the Xerox Award for Faculty Research from UIUC in 2007, and the Young Author Best Paper Award from IEEE in 2008. He has contributed to several tech-transfer efforts, including as the Co-Founder and the CTO of Personify, and the Chief Scientist of Misfit. He was an Associate Editor of the IEEE Transactions on Image Processing.
\end{IEEEbiography}
\vspace{11pt}


\vfill

\end{document}